%% file: acl24_prompt.tex
\pdfoutput=1
\documentclass[11pt]{article}

\usepackage[final]{acl}

\usepackage{times}
\usepackage{latexsym}

\usepackage[T1]{fontenc}
\usepackage[utf8]{inputenc}

\usepackage{microtype}

\usepackage{inconsolata}

\usepackage{graphicx}
\usepackage{amsmath}
\usepackage{amssymb}
\usepackage{booktabs}
\usepackage{subcaption}
\usepackage{caption}
\usepackage{multirow}
\usepackage{bm}
\usepackage{algorithm}
\usepackage{caption}
\usepackage{enumitem}
\usepackage{array}
\usepackage{enumitem}
\usepackage{color}
\usepackage{colortbl}
\usepackage[export]{adjustbox}

\DeclareMathOperator*{\E}{\mathbb{E}}

\title{Prompt Refinement with Image Pivot for Text-to-Image Generation}

\author{
  \textbf{Jingtao Zhan\textsuperscript{1}\thanks{jingtaozhan@gmail.com}},
  \textbf{Qingyao Ai\textsuperscript{1}\thanks{Corresponding author: aiqy@tsinghua.edu.cn}},
  \textbf{Yiqun Liu\textsuperscript{1}},
  \textbf{Yingwei Pan\textsuperscript{2}},
\\
  \textbf{Ting Yao\textsuperscript{2}},
  \textbf{Jiaxin Mao\textsuperscript{3}},
  \textbf{Shaoping Ma\textsuperscript{1}},
  \textbf{Tao Mei\textsuperscript{2}},
\\
\\
  \textsuperscript{1}Department of Computer Science and Technology, Tsinghua University \\
   Zhongguancun Laboratory; Beijing, China \\
  \textsuperscript{2}HiDream.ai; Beijing, China \\
  \textsuperscript{3}Gaoling School of Artificial Intelligence, Renmin University of China; Beijing, China
}

\begin{document}
\maketitle
\input{sec/0_abstract}    
\input{sec/1_intro}

\input{sec/2_related}

\input{sec/3_method}

\input{sec/4_experiment}
\input{sec/5_conclusion}

\input{sec/ethics_impact}

\section*{Acknowledgments}
This work is supported by Quan Cheng Laboratory (Grant No. QCLZD202301).

\bibliography{custom}

\appendix

\input{sec/appendix}

\end{document}

%% file: sec/0_abstract.tex
%\begin{abstract}
%For text-to-image generation, automatically refining user-provided natural language prompts into the keyword-enriched prompts favored by systems is essential for the user experience. Such a prompt refinement process is analogous to translating the prompt from ``user languages'' into ``system languages''. Traditional prompt refinement approaches rely on acquiring high-quality refinement pairs for training; however, the scarcity of such parallel corpora hampers their efficacy. Inspired by zero-shot machine translation techniques, we introduce Prompt Refinement with Image Pivot (PRIP). PRIP innovatively uses the latent representation of a user-preferred image as an intermediary ``pivot'' between the user and system languages. It decomposes the refinement process into two data-rich tasks: inferring representations of user-preferred images from user languages and subsequently translating image representations into system languages. To enhance the process, we embed reinforcement learning for end-to-end optimization and employ residual connections for information preservation. Extensive experiments show that PRIP substantially outperforms various prompt refinement baselines, demonstrating the efficacy of pivoting. Moreover, PRIP not only improves text-to-image systems seen during training but also effectively transfers to an unseen system in a zero-shot manner.
%\end{abstract}
\begin{abstract}
For text-to-image generation, automatically refining user-provided natural language prompts into the keyword-enriched prompts favored by systems is essential for the user experience. Such a prompt refinement process is analogous to translating the prompt from ``user languages'' into ``system languages''. However, the scarcity of such parallel corpora makes it difficult to train a prompt refinement model. Inspired by zero-shot machine translation techniques, we introduce Prompt Refinement with Image Pivot (PRIP). PRIP innovatively uses the latent representation of a user-preferred image as an intermediary ``pivot'' between the user and system languages. It decomposes the refinement process into two data-rich tasks: inferring representations of user-preferred images from user languages and subsequently translating image representations into system languages. Thus, it can leverage abundant data for training. Extensive experiments show that PRIP substantially outperforms a wide range of baselines and effectively transfers to unseen systems in a zero-shot manner\footnote{We have open-sourced code and data at \url{https://github.com/jingtaozhan/PromptReformulate}}.
\end{abstract}

%% file: sec/1_intro.tex
\section{Introduction}
\label{sec:intro}

Recent breakthroughs in text-to-image generation have markedly expanded the boundaries of digital artistry, enabling the creation of visually compelling images with unprecedented ease~\cite{kingma2021variational, ho2020denoising, lu2023specialist, zhu2024SDDiT, zhang2024trip}. However, the complexity of crafting effective prompts presents a significant challenge to average users. This challenge stems from the significant difference between user-provided natural language prompts and the keyword-enriched prompts required for system's high-quality rendering~\cite{brade2023promptify, witteveen2022investigating, parsons2022dalle, chen2023control}.  We term the two kinds of prompts as \emph{user languages} and \emph{system languages}. 
System languages usually include technical terms and artistic references unfamiliar to non-specialists~\cite{liu2022design, oppenlaender2022taxonomy}. Crafting prompts in system languages is not intuitive, even for the system’s developers, and only becomes clear after extensive user experimentation and community-driven insight~\cite{liu2022design, parsons2022dalle, deckers2023manipulating}. 
%Empirically, system languages comprise technical terms and artistic references unfamiliar to non-specialists~\cite{liu2022design, oppenlaender2022taxonomy}, resulting in the mastery of system languages a sophisticated skill beyond everyday users. 
 
Developing a model that automatically refines user languages into system languages is essential to enhance user experience~\cite{openai2023improving, hao2022optimizing, brade2023promptify}. Yet, the shortage of high-quality refinement pairs makes it difficult to train such models. On the one hand, refining prompts requires expertise, which makes annotation expensive. On the other hand, humans can hardly refine prompts to the optimum due to the intricate nature of system languages. As demonstrated in \citet{hao2022optimizing} and our experimental results, human-generated refinement data is sub-optimal for training prompt refinement models.
%refinement models trained with human rewriting data yield compromised generation quality. 
%While previous studies~\cite{hao2022optimizing} attempted to circumvent this by scraping well-performing prompts (deemed as system languages) and synthesizing corresponding user languages, the synthetic distribution of user languages may diverge from real distributions, potentially undermining model performance.

%\begin{figure}
%  \centering
%  \includegraphics[width=0.95\linewidth]{ppt/pivot_illustrate_v2}
%  \caption{Illustration of prompt refinement with and without pivot. Previous approaches~\cite{hao2022optimizing} do not employ a pivot and rely on refinement pairs for training, which is difficult to collect. Our proposed PRIP transforms the refinement task into a two-step process of inferring the user-preferred image followed by decoding the corresponding prompt. Thus, it can leverage abundant training data for both steps and thus significantly enhance the performance.\label{fig:pivot_illustrate}}
%\end{figure}
 
As prompt refinement is analogous to a machine translation task that converts the prompts written in user languages into system languages, the lack of high-quality refinement pairs echoes the challenge of machine translation for low-resource languages~\cite{mhaskarpivot, ranathunga2023neural}, in which a source language is translated into a target language without sufficient parallel corpora for training. MT researchers tackle this through a ``pivoting'' approach~\cite{wu2007pivot, cohn2007machine}, which utilizes a high-resource language as the intermediate ``pivot language''. Thus, the source-target translation is achieved by training two separate models: a source-pivot model and a pivot–target model. Text from the source language is first translated to the pivot language and then to the target language.

Inspired by pivot-based MT solutions, we propose \textbf{P}rompt \textbf{R}efinement with \textbf{I}mage \textbf{P}ivot~(PRIP). PRIP employs the latent representation of a user-preferred image as the pivot between user and system languages. It decomposes prompt refinement into the following two phases. In the first phase, PRIP takes the prompt in user languages as input and infers what images the user prefers. It outputs a latent representation, which focuses on high-level semantics instead of pixel-level details, for the image that is preferred by the user. In the second phase, PRIP takes the latent representation of an image as input and outputs a prompt in the system language that can guide the text-to-image system in rendering this image. By doing so, PRIP reframes a data-limited prompt refinement problem into two data-rich tasks. For example, user-image preference pairs can be constructed using preference simulation models like HPSv2~\cite{wu2023humanv2} or user click behaviors, and image-system decoding pairs can be sampled from interaction logs or prompt-sharing websites. 

%To optimize pivoting, PRIP further integrates two strategies: (1)~An end-to-end reinforcement learning procedure ensures compatibility between the independently trained user-pivot and pivot-system models. (2)~To counter the potential loss of information in the pivot image latent, a residual connection is established between user languages and system languages during inference.

We evaluate PRIP by applying it to various text-to-image models and comparing it with extensive baselines. Results demonstrate that PRIP not only substantially improves the text-to-image system seen during training, but also effectively transfers to various unseen systems in a zero-shot manner. It significantly outperforms a wide range of baselines, including general large language models and prompt refinement models trained with human-generated or synthetic refinement pairs.

% A comprehensive ablation study demonstrates the efficacy of different components of PRIP. 

%% file: sec/2_related.tex
\section{Related Work}

%\noindent \textbf{Text-to-Image Generation}
%Recent breakthroughs in text-to-image generation have markedly expanded the boundaries of digital artistry, enabling the creation of visually compelling images with unprecedented ease~\cite{kingma2021variational, song2020score, ho2020denoising, song2019generative, sohl2015deep}. However, the intricacies of crafting effective prompts present a significant hurdle to the average user. This challenge stems from the stark contrast between casual user prompts, termed as ``user language'', and the more complex ``system language'' that is typically required to produce high-quality imagery. The craft of system language is not intuitive, even for the system’s developers, and only becomes clear after extensive user experimentation and community-driven insight. It often comprises technical terms and artistic references unfamiliar to non-specialists~\cite{liu2022design, oppenlaender2022taxonomy}, rendering the mastery of system language a sophisticated skill beyond everyday users.

\noindent \textbf{Text-to-Image Prompting.}
The automatic refinement of user language into system language is a critical enhancement for a user-friendly text-to-image system~\cite{xie2023prompt}. For a typical text refinement model, training relies heavily on large-scale source-target refinement pairs~\cite{stahlberg2020neural}. However, acquiring such pairs for image prompt refinement is challenging due to the intricate nature of system language, which often exceeds the annotation capacities of crowdsourced workers. To avoid this, some researchers have shifted towards interactive systems that aid users by suggesting enhancements to their prompts, which mitigates but does not dispense with the need for manual refinement~\cite{feng2023promptmagician, brade2023promptify, liu2022design}. Others have attempted to train automatic refinement systems using synthetically generated training pairs, mainly through rephrasing well-crafted prompts into simpler user language forms~\cite{hao2022optimizing}. Yet its synthetic nature often leads to suboptimal refinement performance. Our PRIP addresses the challenge by leveraging the user-pivot-system pipeline, thus avoiding reliance on direct user-system pair annotations.

\noindent \textbf{Prompting Large Language Models (LLMs).} 
There has been research on how to prompt LLMs, such as chain of thoughts~\cite{wei2022chain} and automated template learning~\cite{jiang2020can, haviv2021bertese}. Interested readers can refer to the survey by \citet{liu2023pre}. These studies primarily focus on eliciting the knowledge learned by LLMs during pre-training. However, in cross-modal scenarios like text-to-image generation, the monomodal pre-training of LLMs does not truly capture how prompts influence the image generation process. As a result, generic LLMs do not possess the capability to refine text-to-image prompts, which is also demonstrated in our experiments.

\noindent \textbf{Finetuning Generation Models.}
Several researchers finetune text-to-image models for prompt understanding.
\citet{xu2023imagereward} finetune the generation model on user inputs and use ImageReward as reward.
\citet{zhong2023adapter} finetune the text encoder within the generation model to align simple user inputs with complex prompts.
\citet{deepfloyd_IF_2023} utilize a large language model to process the prompt for better understanding. 
Our experiments show that PRIP can further improve these models.

%% file: sec/3_method.tex
\section{Method}

This section introduces \textbf{P}rompt \textbf{R}efinement with \textbf{I}mage \textbf{P}ivot~(PRIP). 
We first analyze the prompt refinement task.
Then we describe how to decompose the user-system refinement process into user-pivot and pivot-system sub-tasks, where the pivot is the representation of the image preferred by the user.
Finally, we elaborate on the training approaches for the user-pivot-system framework.

\begin{figure*}
  \centering
  \includegraphics[width=1\linewidth]{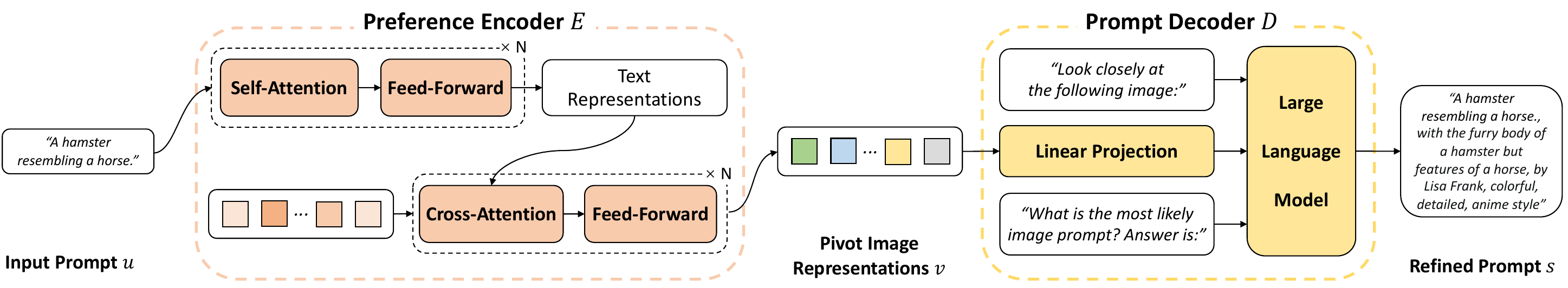}
  \caption{PRIP Model Architecture: a Preference Encoder and a Prompt Decoder. (1) Upon receiving a user prompt, Preference Encoder first applies a transformer to derive a token-level representation. A subsequent transformer leverages cross-attention to deduce the image preference and yields an image representation. (2) Prompt Decoder then employs a linear layer to align the dimensionality. This aligned representation is integrated into a template and input into a large language model, which generates the refined system prompt. \label{fig:prip_model}}
\end{figure*}

\subsection{Problem Analysis}

For text-to-image generation, there exists a notable divergence between the natural language prompts input by non-specialist users, termed \emph{user language}, and the specialized detail-rich prompts, termed \emph{system language}. User languages are often colloquial and ambiguous. System languages include details and artistic terminology, which can guide the systems to yield visually stunning images. 

%User languages are often colloquial and lack the specificity that text-to-image models require to produce high-quality outputs. Conversely, system languages are filled with technical descriptors and artistic terminology that inform the models more effectively.

Prompt refinement systems are optimized to automatically translate user language into system language. We use $\mathcal{R}$ to denote the refinement system, which refines the user language $u$ into the system language $s$ with probability $\mathcal{R}(s|u)$. The text-to-image generation system is denoted as $\mathcal{G}$, where $\mathcal{G}(i|p)$ represents the likelihood of generating an image $i$ from a prompt $p$. The function $f(i,p)$ quantifies the user satisfaction probability correlating a prompt $p$ with image $i$.

Thus, the objective $\mathcal{F}$ of prompt refinement is:
\begin{equation}
\label{eq:refine_objective}
\mathcal{F} = \E_{u,s,i} [f(i,u) \cdot \mathcal{G}(i | s) \cdot \mathcal{R}(s | u)],
\end{equation}
where the refinement system $\mathcal{R}$ should output system language $s$ that maximizes the user satisfaction within the generation probabilities $\mathcal{G}(i | s)$. Ideally, the training of $\mathcal{R}$ would rely on a rich parallel corpus of user-system pairs $\{u, s\}$. However, the rarity of such paired data makes training $\mathcal{R}$ directly with this objective function a considerable challenge.

Viewing $\mathcal{R}$ as a translation model, with $u$ and $s$ as the respective source and target languages, the above challenge is akin to the zero-shot MT problem~\cite{mhaskarpivot, ranathunga2023neural}, where direct source-target translation pairs are absent. Zero-shot MT overcomes this by using a pivot language $v$, which is a high-resource language and ensures conditional independence between the source and target languages~\cite{wu2007pivot, cohn2007machine, bertoldi2008phrase}. Thus, the source language $u$  can be first translated into the pivot language $v$ and then to the target language $s$. $\mathcal{R}(s | u)$ is formally reframed as:
\begin{equation}
\label{eq:pivot_rewrite}
	\begin{aligned}
		 {\mathcal{R}}(s | u) = {\textstyle \sum_{v}} [{D}(s | v) \cdot {E}(v | u)],
	\end{aligned}
\end{equation}
where $D$ and $E$ denote pivot-target and source-pivot translation models, respectively. The pivot language $v$, being high-resource, addresses the data-scarce problem. During inference, to simplify the translation process, the most probable pivot language $v^*$ is selected $v^* = \arg\max_v E(v | u)$, and the translated output $s^*$ is $s^* = \arg\max_s D(s | v^*)$.
%\begin{equation}
%v^* = \arg\max_v E(v | u), \text{ } s^* = \arg\max_s D(s | v^*)
%\end{equation}
In this paper, we adapt this technique to tackle prompt refinement for text-to-image generation.

\subsection{Model Architecture}
\label{sec:model_architecture}

Building upon the principles of zero-shot MT, we introduce \textbf{P}rompt \textbf{R}efinement with \textbf{I}mage \textbf{P}ivot (PRIP). PRIP utilizes the representation of the user-desired image as the pivot $v$ during the prompt refinement process. The refinement workflow, depicted in Figure~\ref{fig:prip_model}, unfolds in two distinct stages: (1) initially, PRIP uses a Preference Encoder $E$ to infer the user's preferred image from the user language prompt. The Preference Encoder adopts a T5-like architecture~\cite{raffel2020exploring}. It first encodes the user language prompt into token-level latent representations and then employs a cross-attention mechanism to produce the pivot image representations. (2) subsequently, PRIP uses a Prompt Decoder $D$ to decode the corresponding system language prompt. The Prompt Decoder is based on a large language model. It accepts the pivot image representations as input, which are then projected to the required dimensionality by a linear layer. The generation process is guided by a prompt template. The Prompt Decoder’s output is the refined, system language prompt.

Both user-pivot and pivot-system stages can leverage extensive training data. The user-pivot stage requires user-image preference data. The data can be easily sourced from click logs or synthesized with user preference simulation models like HPSv2~\cite{wu2023humanv2}. This data may also be annotated, which requires less specialized expertise compared with annotating refinement pairs. The pivot-system stage uses image-prompt pairs. Such data is readily-available from the system's generation logs or from online websites where users share prompts~\cite{stablediffusion_prompts, prompthero}. It does not require annotation and simply relies on the input and output correspondences of the system.

PRIP reframes the refinement objective $\mathcal{F}$ as:
\begin{equation}
\label{eq:prip_objective}
	\begin{aligned}
		 \mathcal{F} = \E_{u,s,i} [ f(i,u) \; {\mathcal{G}}(i | s) \; {\textstyle \sum_{v}} {D}(s | v) {E}(v | u) ]
	\end{aligned}
\end{equation}
To effectively optimize this objective, the training process of PRIP is in two stages. The initial stage involves deriving an approximate objective and employing rich parallel data $\{u,v\}$ and $\{v,s\}$ to warm up PRIP. The subsequent stage leverages reinforcement learning to directly optimize the above objective. They are detailed in the following.

\subsection{Disentangled Supervised Training}

%Rather than directly employing the composite objective as detailed in Eq.~(\ref{eq:prip_objective}), 
We adopt two objectives approximate to Eq.~(\ref{eq:prip_objective}) to warm up PRIP. They enable the use of rich, readily-available data for training. They are derived as:
\begin{equation*}
	\begin{aligned}
		 \mathcal{F} \geq \E_{u,s,i} [f(i,u) \cdot {\mathcal{G}}(i | s) \cdot {D}(s | i) \cdot {E}(i | u)] \\
		 = \; \E_{u,i} \left[ f(i,u) \; {E}(i | u) \right] \cdot \E_{s,i} \left[ {\mathcal{G}}(i | s) \; {D}(s | i) \right] + {\rm Cov}
	\end{aligned}
\end{equation*}
The inequality approaches equality when the evaluated expression is zero for $v \neq i$. This is possible when $\mathcal{G}$ and $D$ form perfect one-to-one correspondences, zeroing $\mathcal{G}(i | s) \cdot D(s | v)$ for all $v \neq i$. For text-to-image systems with strong prompt-following abilities, this simplification of a one-to-one mapping is close and the approximation is reasonable. The term $\rm Cov$ is the covariance between $f(i,u) \cdot E(i | u)$ and $\mathcal{G}(i | s) \cdot D(s | i)$. When they are uncorrelated, the covariance term reduces to zero. We temporarily disregard the covariance and adopt the product of expectations as an approximate surrogate for the original objective. The resulting training objectives are individually focused:
\begin{align}
& \max_{E} \E_{u,i} \; [ f(i,u) \cdot {E}(i | u) ] & \label{eq:train_pref_encoder} \\
& \max_{D} \E_{s,i} \; [ {\mathcal{G}}(i | s) \cdot {D}(s | i) ] & \label{eq:train_prompt_decoder}
\end{align}
These refocused optimization targets permit disentangled training of each module. The subsequent sections detail the specific training processes.

\begin{figure}
  \centering
  \includegraphics[width=0.87\linewidth]{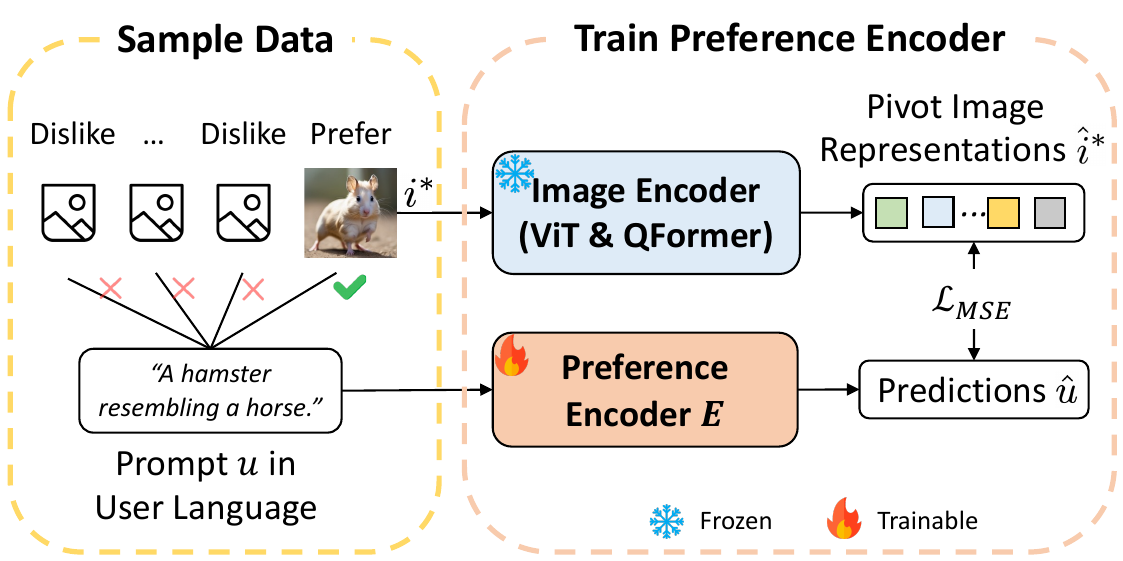}
  \caption{Training Preference Encoder: Prompts and preferred images are paired to create the training set. The objective is to minimize the Mean Squared Error between the ground-truth image representations and the predictions from the Preference Encoder.\label{fig:train_pref_encoder}}
\end{figure}

\subsubsection{Training User-Pivot Preference}

User-pivot transition is guided by Eq.~(\ref{eq:train_pref_encoder}), 
The Preference Encoder is trained to predict the image that can maximize user satisfaction.

As shown in Figure~\ref{fig:train_pref_encoder}, the training process is in two steps. 
(1) Firstly, we sample the image $i^*$ that can maximize user satisfaction $f(i, u)$, namely $i^*=\arg\max_i f(i,u)$. 
Sampling such data is easier than annotating user-system refinement pairs: annotating preference does not require special expertise for annotators and preference can also be extracted from click behaviors in interaction logs. 
%In our experiments, we simulate user preferences through ImageReward~\cite{xu2023imagereward} and HPSv2~\cite{wu2023humanv2} when selecting preferred images $i^*$.
(2) Secondly, The Preference Encoder $E$ is trained to predict $i^*$ given $u$. The user language $u$ is processed by the Preference Encoder, outputting the predicted representation $\hat{u}$. The image $i^*$ is input into an image encoder and results in its semantic representation $\hat{i}^*$. The discrepancy between these representations is minimized with MSE loss:
\begin{equation}
\label{eq:train_pref_encoder_loss}
	\mathcal{L}_{MSE} = || \hat{i}^* - \hat{u} ||^2_2 
\end{equation}
With this training process, the Preference Encoder aligns with user preferences by learning to imagine the user-preferred images.

%This training protocol aligns the pivot's latent space with that of the image encoder. By utilizing a fixed image encoder for both the Preference Encoder and the upcoming Prompt Decoder training, we ensure that the pivot representation space remains consistent across user-pivot and pivot-system transitions. Such consistency is critical for the seamless integration within the PRIP framework, which will be further detailed in the subsequent subsection.

\subsubsection{Training Pivot-System Decoding}
\label{sec:decoder_training}

\begin{figure}
  \centering
  \includegraphics[width=1\linewidth]{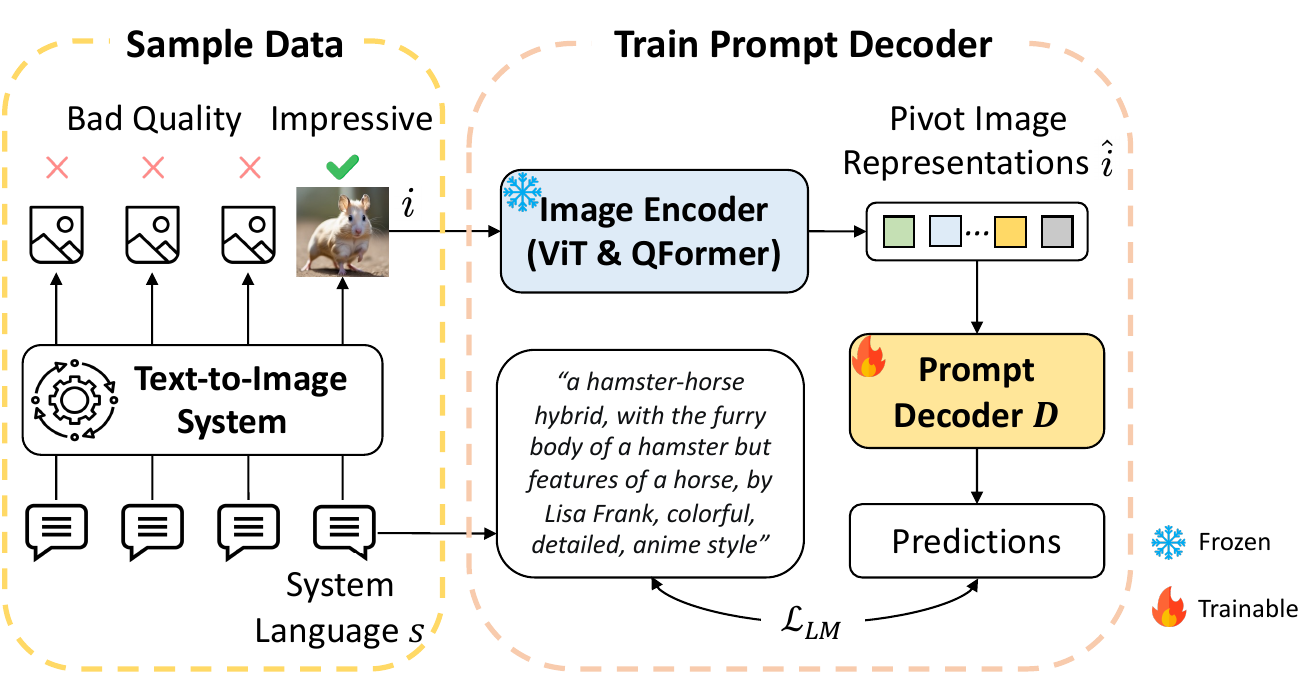}
  \caption{Training Prompt Decoder: Prompts that can generate impressive images are sampled as the system language. The objective is to predict the system language based on the associated image representation.\label{fig:train_prompt_decoder}}
\end{figure}

Pivot-system transition is guided by Eq.~(\ref{eq:train_prompt_decoder}). The Prompt Decoder is trained to reconstruct the system language from the pivot image representation. 

As illustrated in Figure~\ref{fig:train_prompt_decoder}, the training process is in two steps.
(1) Firstly, we collect system language prompts $s$ and their corresponding generated images $i$. The system languages are high-quality prompts suitable for the generation system. They can be sampled from the user-submitted prompts in the generation log or from websites where users share well-performing prompts.
%We collect data under the premise that user-submitted prompts can serve as close approximations of system languages when the generated images exhibit high relevance, strong user preference, and aesthetic appeal. This allows us to harvest a wealth of prompt-image pairs from the interaction log, providing an adequate dataset for training.
(2) The Prompt Decoder is trained to generate system language $s$ from the image $i$. A frozen image encoder processes the image $i$, creating its representation $\hat{i}$. This representation serves as the training context for the Prompt Decoder to predict the system language. Autoregressive language modeling objective is used and is formulated as follows:
\begin{equation}
\label{eq:train_prompt_decoder_loss}
	\mathcal{L}_{LM} = - {\textstyle \sum_n} {\rm log} \; {D}(s_n | s_{1:n-1}, \hat{i}),
\end{equation}
where $s_n$ is the $n$-th token of $s$.
In this way, the Prompt Decoder aligns with the generation systems by learning to reverse the generation process. 

%To integrate the user-pivot and pivot-system processes, we employ the same frozen image encoder for both training stages. This ensures that the latent representation space output from the user-pivot is in direct correspondence with the pivot-system's input space, paving the way for a seamless user-pivot-system transition within the PRIP framework.

\subsection{End-to-End User-Pivot-System Training}
\label{sec:end_to_end_rl}

While the previously described training stages effectively optimize the user-pivot and pivot-system transitions on supervised data, they serve primarily as approximations of the ultimate objective presented in Eq.~(\ref{eq:prip_objective}). To bridge this gap, we leverage the previous training as a warm-up stage and subsequently employ Eq.~(\ref{eq:prip_objective}) for end-to-end training.

\begin{figure}
  \centering
  \includegraphics[width=1\linewidth]{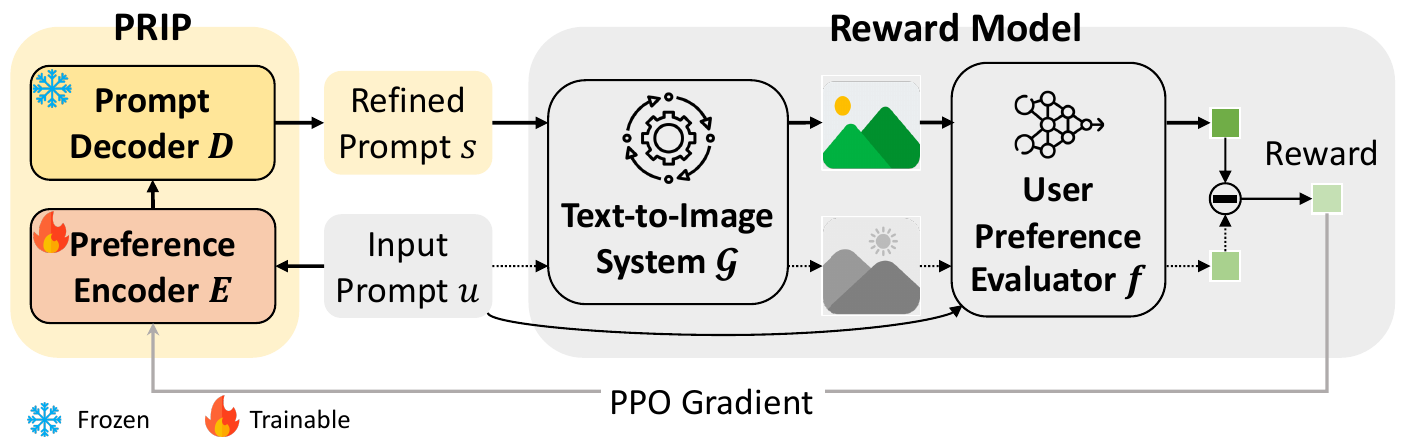}
  \caption{End-to-End RL Training: Given a user prompt, PRIP generates a refined prompt, and Reward Model evaluates user preference scores for generated images. The differential in scores serves as the reward, and PRIP is updated with PPO Gradient.
  \label{fig:reinforce_learn}}
\end{figure}

We adopt reinforcement learning (RL) and define the reward as the differential in preference scores. This is equivalent to Eq.~(\ref{eq:prip_objective}):
\begin{equation*}
	\begin{aligned}
		 \arg \max_{D,E} \mathcal{F} = \arg \max_{D,E} \E \; [ {\rm reward} \; {D}(s | v) {E}(v | u) ], \\
		 {\rm reward} = {\textstyle \sum_i} f(i,u) {\mathcal{G}}(i | s) - {\textstyle \sum_{i'}} f(i',u) {\mathcal{G}}(i' | u)
	\end{aligned}
\end{equation*}
The training workflow is depicted in Figure~\ref{fig:reinforce_learn}. For any given prompt, we generate two image sets: one set from the initial prompt and another from the refined system prompt. The reward is computed as the differential in preference scores between these two sets. To enhance training efficiency, Prompt Decoder remains frozen due to the significant computational costs of training a large language model. Only Preference Encoder is updated with proximal policy optimization (PPO)~\cite{schulman2017proximal}, a well-regarded RL algorithm.

\subsection{Inference Process}

During inference, the Preference Encoder and the Prompt Decoder are concatenated as a user-pivot-system pipeline. 
Given a prompt, the Preference Encoder predicts the representation of the user-preferred image. Then, the Prompt Decoder decodes the representation and outputs the refined prompt. 
The refined prompt is input to the generation system for image rendering. 

Furthermore, since the Prompt Decoder cannot directly access the initial prompt, it might result in a hallucination problem~\cite{ji2023survey} when the refined prompt is generated. A solution is to provide the initial prompt as additional context to the Prompt Decoder during inference. This paper uses a straightforward approach: prepending the initial prompt as a prefix and using the Prompt Decoder for expansion. The refined prompt starts with the initial prompt and consists of many details added by the Prompt Decoder.
Future studies may investigate other methods. For example, constrained beam search can also use the initial prompts as additional context by constraining the refined prompts to be semantically close to the initial prompts. We leave this exploration to future work.

%% file: sec/4_experiment.tex
\section{Experimental Setup}

\subsection{Evaluation Setup}
\label{sec:evaluate_setup}

\noindent
\textbf{Generation Systems.}
We evaluate the prompt refinement performance for a wide range of generation systems. 
When PRIP and other refinement baselines are trained, they are optimized for \textbf{Stable Diffusion 1.4~(SD1.4)}~\cite{rombach2022high}. Thus, the refinement performance on SD1.4 corresponds to the in-distribution performance.
We also employ various advanced generation systems to evaluate the out-of-distribution performance. They are:
(1) \textbf{Stable Diffusion XL base 1.0~(SDXL)}~\cite{podell2023sdxl}, a state-of-the-art generation model that is much stronger than SD1.4.
(2) \textbf{Deepfloyd-IF~(IF)}~\cite{deepfloyd_IF_2023}, an advanced model for high degree of prompt understanding. It employs a T5-XXL model for prompt processing. 
(3) \textbf{SUR}~\cite{zhong2023adapter}, which is specifically proposed to understand simple user inputs for high-quality rendering. 
(4) \textbf{ReFL}~\cite{xu2023imagereward}, which is initialized from SD1.4 and further trained with the same reward model as PRIP. The difference is that ReFL trains the generation model while PRIP modifies the input.
We also attempted to evaluate on DALL-E 3~\cite{openai2023improving}. However, due to restrictions on its inputs, such as not allowing artists' names, our test prompts and refined prompts often do not meet these restrictions and make evaluation impossible.

\noindent
\textbf{Dataset.} 
We conduct evaluations using the HPS prompt dataset~\cite{wu2023humanv2}, which includes a wide variety of prompts mined from user interactions and image captions.
It is a standard benchmark to evaluate text-to-image generation performance~\cite{clark2023directly, wallace2023diffusion}.
The prompts are categorized into Anime, ConceptArt, Painting, and Photo. Each category contains 800 prompts. For each prompt, we generate four images to ensure a robust assessment.

%\noindent
%\textbf{Refinement Setup.}
%Since hallucination problem of language models~\cite{ji2023survey} can result in inconsistent semantics between the initial and the refined prompts, we utilize a straightforward counter measure during inference. 
%When generating the refined prompt, we prepend the initial prompt as a prefix and uses the model for expansion. Thus, the refined prompt always starts with the initial prompt, and the refinement models should expand different modifer words depending on this context.
%It effectively addresses the hallucination problem.
%Future studies can investigate more effective decoding methods, such as constrained beam search.
 
\noindent
\textbf{Metrics.}
We employ both automated and human judgment. Automated assessments utilize ImageReward~\cite{xu2023imagereward} and HPSv2~\cite{wu2023humanv2}, which are trained to mimic human preferences and have been demonstrated to accurately align with actual human judgments. 
Humans annotate relevance and win ratios. Relevance is the prompt-image alignment on a scale of 0 (irrelevant) to 2 (highly relevant). Win ratio (and tie) shows pairwise prompt-image preference between two generation systems. We randomly sample $30$ prompts per category and report the annotation results averaged on all $120$ prompts. Each pair is annotated by three people who are familiar with text-to-image generation and of different backgrounds.

%During inference, we prepend the initial prompt as a prefix and uses the prompt refinenemt model to expand the prompt as the refined prompt. Using initial prompts as prefix is due to the characteristics of this dataset: most evaluation prompts are concise and clear descriptions and do not need to be reformulated. Using them as prefix can avoid hallucination problems of language models. 

\input{tables/compare_on_sd14}

\subsection{Baselines}

We compare PRIP against a comprehensive set of prompt refinement baselines: 
(1) \textbf{GPT3.5 \& 4}: They are generic language models and are not tailored for prompt refinement. To guide them, we use a popular prompt template from \citet{bluelovers2023chatgpt} and slightly modify it for this task. The template contains guidance and examples. 
(2) \textbf{PromptistSFT} \& \textbf{PromptistRL}~\cite{hao2022optimizing}: PromptistSFT is trained on synthesized parallel data: system languages are collected from prompt-sharing websites, while user languages are rephrased from the system languages to simple form by ChatGPT. Initialized from PromptistSFT, PromptistRL undergoes an RL process with CLIP~\cite{radford2021learning} and Aesthetic scores~\cite{aesthetic_predictor} as reward.
(3) \textbf{Rew-Syn} \& \textbf{Rew-Syn+RL}: We enhance PromptistSFT and PromptistRL by aligning them with user preference. Rew-Syn uses ImageReward and HPSv2 to filter out the PromptistSFT training pairs whose refinement does not improve satisfaction scores. Rew-Syn+RL utilizes ImageReward and HPSv2 as reward, which is identical to PRIP's. 
(4) \textbf{Rew-Log} \& \textbf{Rew-Log+RL}: Rew-Log extracts human rewriting pairs from a large-scale interaction log~\cite{wang2022diffusiondb} by pairing the first and the last prompts in the same session. It filters out the pairs that do not improve ImageReward or HPSv2 scores. Rew-Log+RL is initialized from Rew-Log and undergoes the same RL training process as PRIP's.

%The following refinement baselines are initialized with the above checkpoints and further trained with RL. (1) \textbf{PromptistRL}~\cite{hao2022optimizing}: Starting with PromptistSFT, this model employs CLIP~\cite{radford2021learning} and Aesthetic scores~\cite{aesthetic_predictor} as a reward signal during RL training. (2 \& 3) \textbf{Rew-Syn+RL \& Rew-Log+RL}: These models, respectively initialized with Rew-Syn and Rew-Log, undergo further training with an RL recipe identical to PRIP's.

\subsection{Implementation Details}
\label{sec:implement_details}

\textbf{Architecture.}
The model architecture is as follows. Preference Encoder and Prompt Decoder are initialized with FLAN-T5-Large~\cite{chung2022scaling} and Llama2-7B~\cite{touvron2023llama}, respectively. We use the vision component of BLIP-2~\cite{li2023blip2} as the frozen Image Encoder.

\noindent
\textbf{Data.}
Training data for PRIP can be easily acquired, as discussed in Section~\ref{sec:model_architecture}. In this paper, we collect data from DiffusionDB~\cite{wang2022diffusiondb}, a real interaction log between 10k users and SD1.4. 
Since this dataset logs the multiple generated images for each prompt, we can sample the most-preferred image for training the user-pivot model (the Preference Encoder). We use ImageReward to simulate user preference and select the highest-scored image as the pivot. 
We also observe that there exist high-quality prompts in the log that can serve as the system languages. 
Therefore, we sample these prompts and the associated images for training the Prompt Decoder model. The sampling criterion is empirically set: the CLIP and Aesthetic scores are above $0.28$ and $5.2$, respectively.
Future studies can explore other data resources.

Please refer to Appendix~\ref{sec:computational_exp} for more details.
%For user-pivot training, we first sample prompt-image preference pairs based on thresholds for CLIP, Aesthetic, ImageReward, and HPSv2 scores, which are empirically set to $0.28$, $5.5$, $1$, and $0.27$. Then the image with the highest ImageReward score is selected for each prompt, resulting in a dataset of $325{\rm k}$ pairs. Preference Encoder is trained for $3$ epochs using Adafactor~\cite{shazeer2018adafactor} with a learning rate of $0.001$.
%
%For pivot-system training, we collect system languages by sampling prompts whose mean CLIP, max CLIP, and mean Aesthetic scores are above thresholds, empirically set to $0.24$, $0.28$, and $5.2$, respectively. The highest CLIP-scored image is chosen for each prompt, culminating in $985{\rm k}$ prompt-image pairs. Prompt decoder is trained for $2$ epochs with AdamW~\cite{kingma2014adam} and a learning rate of $2\times 10^{-5}$.
%
%In user-pivot-system RL training, we generate four images per prompt, with ImageReward and HPSv2 to output preference scores. To ensure a balance of quality and diversity, we set the sampling temperature to $1.0$. RL is performed for $1,000$ steps with a batch size of $512$, employing the Adafactor optimizer at a constant learning rate of $0.001$.

\input{tables/compare_on_sdxl}

\input{tables/prompt_cases_sdxl_images}

\section{Experimental Results}

This section presents the experimental results. 
We first evaluate models on various generation systems, including the one used during training (in-distribution scenario) and several unseen, advanced systems (out-of-distribution scenario).
Then, we show how PRIP refines the prompt by presenting several cases.
Finally, a comprehensive ablation study demonstrates the effectiveness of pivoting.

\subsection{In-Distribution Performance}

Table~\ref{tab:compare_on_sd14} presents the in-distribution refinement performance. We have the following observations: (1) Results indicate that generic language models like GPT3.5 and GPT4 do not excel at refining image prompts. We find that GPT3.5 merely rephrases prompts without adding details and sometimes hallucinates, which results in its output being mostly ineffective. As for GPT4, it can effectively add rich details compared with GPT3.5. Yet its added details often misalign with the initial prompts, leading to even worse performance. (2) Synthetically generated refinement pairs also demonstrate limited efficacy, with neither PromptistSFT nor Rew-Syn surpassing the SD1.4 baseline. The user languages are synthesized by rephrasing high-quality prompts. Yet such synthesized user inputs are different from real inputs and compromise the model performance.
%(3) In contrast, Rew-Log, which leverages real human rewriting data extracted from the interaction log, emerges as the strongest non-RL baseline.  
(3) In contrast, PRIP does not rely on any synthesized or human-generated refinement pairs that are usually low-quality and noisy. Results demonstrate that PRIP significantly outperforms all baselines. 

%PRIP$\backslash$RL, trained without any user-system refinement pairs, outperforms these non-RL baselines. RL training further amplifies PRIP advantage, underscoring the limitations of baselines due to lack of high-quality refinement pairs and affirming PRIP’s robust solution to this issue.

\subsection{Out-of-Distribution Performance}

Table~\ref{tab:compare_on_sdxl} shows the performance on four state-of-the-art generation systems that are unseen during training.
As introduced in Section~\ref{sec:evaluate_setup}, these systems employ special techniques to improve prompt-understanding abilities. For example, IF uses a large language model to process prompts, and ReFL is finetuned on user inputs with preference feedback. 
Even on these systems, PRIP still presents significant improvements.
Furthermore, according to the human annotation results in Table~\ref{tab:compare_on_sd14} and \ref{tab:compare_on_sdxl}, PRIP's performance improvement is more pronounced on SDXL than SD1.4. We closely examine the output and find that SD1.4 sometimes struggles to process the rich details added by PRIP while SDXL can.
The results indicate that PRIP is even more suitable for advanced generation systems. It also implies that existing generation systems all prefer prompts with rich details and professional terms.
In Appendix~\ref{sec:compare_with_refl}, we provide an additional analysis showing how PRIP leads to robust improvement. 

%PRIP exhibits remarkable adaptability to SDXL, substantially enhancing image quality, in contrast to the baselines which show minimal improvement. Notably, PRIP's performance improvement is more pronounced on SDXL than SD1.4. For instance, PRIP$\backslash$RL's enhancements are observable in the Photo category for SDXL but not for SD1.4, and PRIP achieves higher win ratios on SDXL. A close examination at the output reveals that while SD1.4 sometimes struggles to integrate the original user input with PRIP's detailed augmentations, SDXL handles the combination adeptly, facilitating more consistent enhancements from PRIP.

\subsection{Case Studies}

Table~\ref{tab:prompt_case_study} presents several refinement examples. We can see that PRIP expands user inputs with details and stylistic elements. The details are rich, professional, and tailored for each user input.
For example, in the first case, PRIP adds the monkey's wearing, action, and environment. It also specifies that the image is a tourism pamphlet cover. These make the rendered image both closely relevant to the user input and aesthetically-pleasing.
Moreover, some added terms are professional artist names and beyond the capability of average users, such as ``Thomas Kinkade'' in the third case, ``Pauline Baynes'' in the fourth case, and ``Lisa Frank'' in the seventh case. With PRIP automatically adding these professional terms, the text-to-image systems can become more user-friendly.

%PRIP not only adds intricate details to the prompts (as in the first example, which includes attributes like the subject's attire and action) but also adeptly tailors the style to match the input prompt (specifying a tourism pamphlet cover style in the first case). These instances illustrate PRIP's effectiveness in converting user languages into detailed and contextually rich system languages.

%We can see that 1) PRIP enriches the prompts with nuanced details. In the first case, it adds attributes related to the monkey's attire, expression, action, and identity. 2) PRIP tailors the style of the generated images to align with the input prompts. In the first case, it specifies the image to emulate a tourism pamphlet cover. These examples show PRIP's efficacy in transforming user input into detailed and contextually rich prompts.

\subsection{Ablation Study}

\input{tables/ablation_study}

To evaluate the contributions of PRIP's components, we perform a detailed ablation study, examining the exclusion of the following elements: 
(1) $\backslash$User-Pivot Preference: For user-pivot training as in Eq.~(\ref{eq:train_pref_encoder_loss}), the ground truth is a random image generated from this prompt instead of the image with the highest preference score. This investigates whether the pivot should be a user-preferred image.
(2) $\backslash$Pivot-System Decoding: In pivot-system training as in Eq.~(\ref{eq:train_prompt_decoder_loss}), Prompt Decoder is not provided with image as input and is trained using a basic language modeling loss. The trained model is familiar with prompts but not capable of decoding image pivots.
(3) Without RL (PRIP$\backslash$RL): We evaluate the PRIP model that does not undergo an end-to-end RL process, as presented in Section~\ref{sec:end_to_end_rl}. 
%(3) $\backslash$User-System Residual: During prompt decoding in Eq.~(\ref{eq:user_system_residual}), user input does not serve as a prefix.

Table~\ref{tab:ablation_study} presents the ablation results. It demonstrates that all three components are vital to PRIP.
According to the performance of $\backslash$Pivot-System Decoding, pivot-system training is critical. Without it, the performance substantially degenerates.
This indicates that PRIP relies on this process to learn to decode the image pivot. 
``$\backslash$User-Pivot Preference'' replaces the user-preferred image with a random image, which also results in a performance drop. This demonstrates the importance of aligning user preference by using the best image as the pivot during training.
RL can substantially improve the effectiveness of PRIP. 
Yet, it still relies on the user-pivot and pivot-system to provide a good warmup process. This is in line with the observations by \citet{zheng2023secrets} that the exploration space of the language model is too large and convergence of RL is formidable without a good start point. 

%the critical role of the user-pivot and pivot-system training stages. Without either, relying solely on RL for user-pivot-system training significantly underperforms, which is inline with the observations by \citet{zheng2023secrets} that the exploration space of language model is too large and convergence of RL is formidable without a good starting point. 

%Results also highlight the vital function of the user-system residual connection. Its absence leads to a significant performance drop, indicating that it effectively counters the error-prorogation problem via providing a parallel pathway to the user-pivot-system pipeline.

%caused by the inaccuracy in pivot prediction within the user-pivot-system pipeline.

%, ensuring the integrity and efficacy of the overall refinement process.
 
%It is worth noting that using supervised finetuning process for warm up is essential to the efficacy of RL. Although RL facilitates directly optimizing the target objective, the search space is too large and convergence is formidable without a good starting point. This is demonstrated by \cite{zheng2023secrets} and in our ablation studies.

\subsection{Scaling Analysis}

We investigate how the model size affects PRIP performance.
We use two smaller models, TinyLlama with 1.1B parameters~\cite{zhang2024tinyllama} and GPT2-Large with 0.78B parameters~\cite{radford2019language}. Although TinyLlama has fewer parameters than Llama2-7B~\cite{touvron2023llama}, it was trained on 2.5T tokens, compared to 2T tokens for the latter. Thus, TinyLlama is a strong ``small'' model, while GPT2-Large is a relatively weaker ``small'' model.

The setup is as follows. We use both models to initialize the Prompt Decoder. We do not use the RL training process as described in Section~\ref{sec:end_to_end_rl} to save cost. We only train the models with pivot-system pairs as shown in Section~\ref{sec:decoder_training}. We pair the Preference Encoder with these two new Prompt Decoders to form two new PRIP models. We test the prompt refinement performance on the Anime and Painting datasets using the SD1.4 model for image generation.

\input{tables/scaling.tex}

Based on the results in the table, we can observe:
(1) As the capability of the base model increases, the model's prompt refinement ability also improves gradually. A more capable base model helps PRIP better infer the system prompt language from the pivot image representation, thus achieving better image generation results.
(2) TinyLlama demonstrates a substantial improvement over GPT2-Large in our task, suggesting that the extensive pre-training contributes to the superior performance in this downstream task.
(3) We also observed that the performance of TinyLlama is approaching that of Llama2-7B, even though their parameter scales differ by a factor of 7. This suggests that with the help of PRIP's extensive training data, the demand for the size of the base model has become smaller. A well-optimized small model can achieve performance close to that of a large model.

%% file: tables/compare_on_sd14.tex
\newcommand{\B}{\bfseries}

\newcommand{\sig}{$^{*}$}
\newcommand{\emp}{\,\;}

\begin{table*}[ht]
\centering
\small
\scalebox{0.8}[0.8]{
\begin{tabular}{l|cccc|cccc|ccc}
   	\toprule
%    \multirow{2}{*}{Method} 
    Evaluation Metric & \multicolumn{4}{c|}{ImageReward} & \multicolumn{4}{c|}{HPSv2} & Relevance & Win & Win+Tie\\
    Dataset & Anime & ConceptArt & Painting & Photo & Anime & ConceptArt & Painting & Photo & All & All & All \\
    \midrule
SD1.4                          & 0.038\sig & 0.185\sig & 0.190\sig & 0.130\sig & 27.42\sig & 26.86\sig & 26.86\sig & 27.57\sig &  1.38\emp & 1\%\sig & \B 100\%\emp \\
\midrule
+ GPT3.5                       & -0.037\sig & 0.030\sig & 0.126\sig & -0.005\sig & 27.36\sig & 26.77\sig & 26.87\sig & 27.41\sig &  -- & -- & -- \\
+ GPT4                         & -0.143\sig & -0.024\sig & 0.030\sig & -0.196\sig & 27.29\sig & 26.71\sig & 26.76\sig & 27.28\sig &  -- & -- & -- \\
+ PromptistSFT                 & -0.140\sig & -0.083\sig & 0.010\sig & -0.287\sig & 27.19\sig & 26.60\sig & 26.77\sig & 26.88\sig &  -- & -- & -- \\
+ Rew-Syn                      & -0.015\sig & 0.056\sig & 0.223\sig & -0.221\sig & 27.35\sig & 26.79\sig & 26.99\sig & 26.96\sig &  -- & -- & -- \\
+ Rew-Log                      & 0.066\sig & 0.151\sig & 0.173\sig & 0.063\sig & 27.44\sig & 26.87\sig & 26.86\sig & 27.51\sig &  -- & -- & -- \\
+ PromptistRL                  & -0.009\sig & 0.092\sig & 0.211\sig & -0.060\sig & 27.29\sig & 26.73\sig & 26.89\sig & 26.97\sig &  1.20\sig & 38\%\sig & 84\%\emp \\
+ Rew-Syn+RL                   & 0.079\sig & 0.135\sig & 0.246\sig & 0.138\sig & 27.46\sig & 26.85\sig & 26.99\sig & 27.70\sig &  1.31\sig & 40\%\sig & 89\%\emp \\
+ Rew-Log+RL                   & 0.028\sig & 0.177\sig & 0.187\sig & 0.105\sig & 27.42\sig & 26.85\sig & 26.86\sig & 27.55\sig &  1.33\sig & 6\%\sig & 96\%\emp \\
+ \textbf{PRIP}                         & \B 0.346\emp & \B 0.443\emp & \B 0.576\emp & \B 0.252\emp & \B 27.97\emp & \B 27.45\emp & \B 27.65\emp & \B 28.03\emp & \B  1.45\emp & \B 73\%\emp & 94\%\emp \\
    \bottomrule
\end{tabular}
}
\caption{In-distribution refinement performance on the seen system~(SD1.4). Win/Tie ratio shows preference against SD1.4, and SD1.4's ``Win+Tie'' is always 100\%. Human annotation is only conducted on RL-based methods to save costs. $^*$ indicates PRIP significantly outperforms the baseline with p-value $< 0.01$ measured by T-Test. PRIP substantially outperforms baselines.}
\label{tab:compare_on_sd14}
\end{table*}

\let\B\undefined
\let\sig\undefined
\let\emp\undefined

%% file: tables/compare_on_sdxl.tex
\newcommand{\B}{\bfseries}

\newcommand{\sig}{$^{*}$}
\newcommand{\emp}{\,\;}

\begin{table*}[ht]
\centering
\small
\setlength{\tabcolsep}{3mm}
\scalebox{0.8}[0.8]{
\begin{tabular}{l|cccc|cccc|ccc}
   	\toprule
%    \multirow{2}{*}{Method} 
    Evaluation Metric & \multicolumn{4}{c|}{ImageReward} & \multicolumn{4}{c|}{HPSv2} & Relevance & Win & Win+Tie \\
    Generation Model & SDXL & IF & SUR & ReFL & SDXL & IF & SUR & ReFL & SDXL & SDXL & SDXL \\
    \midrule
w$\backslash$ o refine                     & 0.866\sig & 0.624\sig & 0.596\sig & 0.421\sig & 27.76\sig & 27.63\sig & 27.82\sig & 27.64\sig &  1.67\emp & 1\%\sig & \B 100\%\emp \\
\midrule
+ GPT3.5                       & 0.753\sig & 0.569\sig & 0.431\sig & 0.316\sig & 27.67\sig & 27.57\sig & 27.72\sig & 27.57\sig &  -- & -- & -- \\
+ GPT4                         & 0.702\sig & 0.455\sig & 0.360\sig & 0.214\sig & 27.67\sig & 27.41\sig & 27.67\sig & 27.49\sig &  -- & -- & -- \\
+ PromptistSFT                 & 0.679\sig & 0.298\sig & 0.374\sig & 0.229\sig & 27.54\sig & 26.93\sig & 27.53\sig & 27.44\sig &  -- & -- & -- \\
+ Rew-Syn                      & 0.817\sig & 0.464\sig & 0.513\sig & 0.347\sig & 27.72\sig & 27.17\sig & 27.67\sig & 27.60\sig &  -- & -- & -- \\
+ Rew-Log                      & 0.850\sig & 0.591\sig & 0.561\sig & 0.401\sig & 27.75\sig & 27.57\sig & 27.79\sig & 27.62\sig &  -- & -- & -- \\
+ PromptistRL                  & 0.833\sig & 0.509\sig & 0.547\sig & 0.404\sig & 27.67\sig & 27.21\sig & 27.65\sig & 27.56\sig &  1.52\sig & 40\%\sig & 88\%\emp \\
+ Rew-Syn+RL                   & 0.874\sig & 0.579\sig & 0.596\sig & 0.459\sig & 27.81\sig & 27.46\sig & 27.83\sig & 27.77\sig &  1.62\emp & 51\%\sig & 89\%\emp \\
+ Rew-Log+RL                   & 0.861\sig & 0.619\sig & 0.573\sig & 0.406\sig & 27.77\sig & 27.60\sig & 27.79\sig & 27.63\sig &  1.67\emp & 4\%\sig & 98\%\emp \\
+ \textbf{PRIP}                         & \B 0.983\emp & \B 0.741\emp & \B 0.789\emp & \B 0.640\emp & \B 28.15\emp & \B 27.90\emp & \B 28.22\emp & \B 28.14\emp &  \B 1.68\emp & \B 82\%\emp & 96\%\emp \\
%w/o refine                     & 0.992  & 0.903  & 0.907  & 0.661  & 28.37  & 27.46  & 27.52  & 27.71 & 1.67 & 0\% & \B 100\% \\
%\midrule
%+ GPT3.5                       & 0.883  & 0.762  & 0.832  & 0.534  & 28.26  & 27.32  & 27.50   & 27.59 & -- & -- & -- \\
%+ GPT4                         & 0.831  & 0.743  & 0.794  & 0.438  & 28.28  & 27.34  & 27.43  & 27.63 & -- & -- & --  \\
%+ PromptistSFT                 & 0.785  & 0.638  & 0.688  & 0.606  & 28.12  & 27.19  & 27.34  & 27.53 & -- & -- & -- \\
%+ Rew-Syn                      & 0.919  & 0.803  & 0.910   & 0.636  & 28.30   & 27.39  & 27.57  & 27.61 & -- & -- & --  \\
%+ Rew-Log                      & 0.972  & 0.880   & 0.897  & 0.652  & 28.33  & 27.46  & 27.50   & 27.73 & -- & -- & -- \\
%+ PromptistRL                  & 0.919  & 0.835  & 0.898  & 0.680   & 28.25  & 27.41  & 27.51  & 27.52 & 1.53 & 40\% & 88\% \\
%+ Rew-Syn+RL                   & 0.997  & 0.873  & 0.941  & 0.685  & 28.38  & 27.45  & 27.61  & 27.81 & 1.62 & 51\% & 89\% \\
%+ Rew-Log+RL                   & 0.987  & 0.897  & 0.903  & 0.655  & 28.37  & 27.47  & 27.52  & 27.71 & 1.67 & 4\% & 98\% \\
%+ \textbf{PRIP}                          & \B 1.085  & \B 1.014  & \B 1.116  & \B 0.717  & \B 28.62  & \B 27.81  & \B 27.98  & \B 28.20 & \B 1.68 & \B 83\% & 98\%  \\
    \bottomrule
\end{tabular}}
\caption{Out-of-distribution refinement performance on unseen systems. Results are averaged on four categories. Human annotation is only conducted on RL-based methods for SDXL to save costs. Win/Tie ratio shows preference against generation without refinement. Note that the first row is generation without refinement and its ``Win+Tie'' is 100\%. 
$^*$ indicates PRIP significantly outperforms the baseline with p-value $< 0.01$ measured by T-Test. PRIP effectively transfers to unseen systems.}
\label{tab:compare_on_sdxl}
\end{table*}

\let\B\undefined
\let\sig\undefined
\let\emp\undefined

%% file: tables/prompt_cases_sdxl_images.tex
\newlength{\myimagewidth}
\setlength{\myimagewidth}{0.17\linewidth} % 设置您希望的全局宽度

\begin{table}[t]
\centering
\scriptsize
\setlength{\tabcolsep}{1pt}
\scalebox{0.9}[0.9]{
\begin{tabular}{m{0.74\linewidth}cc}
\toprule
\textbf{Input Prompt} & \textbf{User} & \textbf{PRIP}  \\ 
\midrule
\parbox[t]{\linewidth}{
\textbf{User}\,\,: A monkey is pictured acting as a DJ. \\
\textbf{PRIP}: A monkey is pictured acting as a DJ., he is wearing headphones and has a large smile on his face, he is holding a record, he is at a rave, he is on the cover of a tourism pamphlet for Florida, he is a resident DJ at ...% a local nightclub
}
& 
\raisebox{-.85\height}{\includegraphics[width=\myimagewidth]{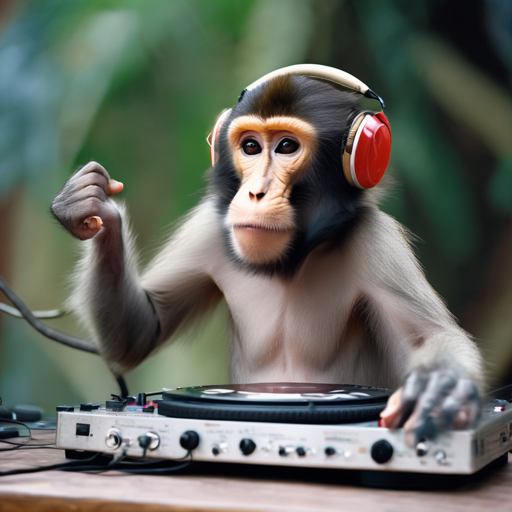}}  &
\raisebox{-.85\height}{\includegraphics[width=\myimagewidth]{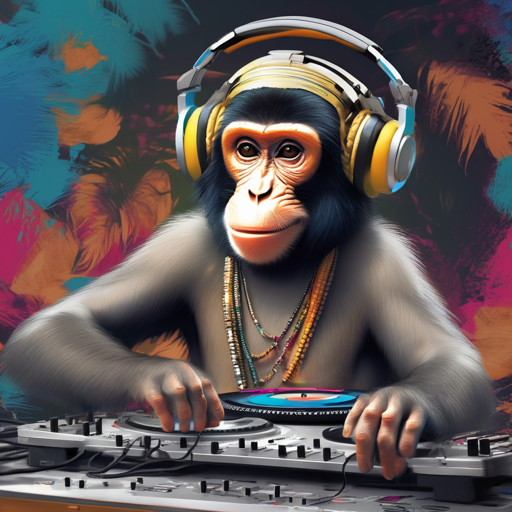}} \\
\midrule
\parbox[t]{\linewidth}{
\textbf{User}\,\,: A Walter White funko pop figurine. \\
\textbf{PRIP}: A Walter White funko pop figurine., intricate, highly detailed, photorealistic, 4k, HDR, smooth, sharp focus, high resolution, award-winning photo, taken at the 2022 EPCOT International Flower and Food Festival} & 
\raisebox{-.85\height}{\includegraphics[width=\myimagewidth]{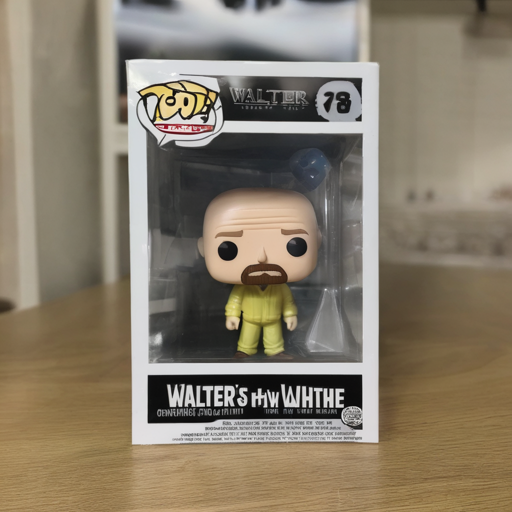}}  &
\raisebox{-.85\height}{\includegraphics[width=\myimagewidth]{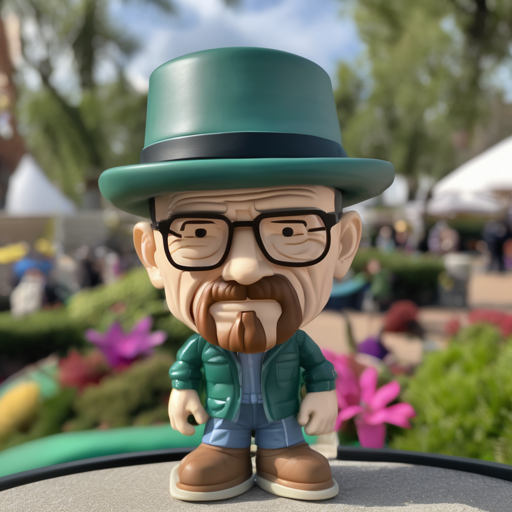}} \\
\midrule
\parbox[t]{\linewidth}{
\textbf{User}\,\,: A mushroom house in a dark forest, with warm light emitting from its windows. \\
\textbf{PRIP}: A mushroom house in a dark forest, with warm light emitting from its windows., by Thomas Kinkade, colorful, vibrant, intricate, highly detailed, deviantart, ... % artstation, by Lisa Frank, by Artgerm
} & 
\raisebox{-.85\height}{\includegraphics[width=\myimagewidth]{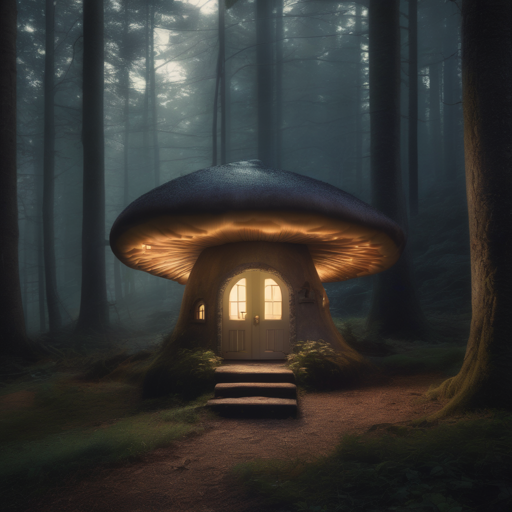}}  &
\raisebox{-.85\height}{\includegraphics[width=\myimagewidth]{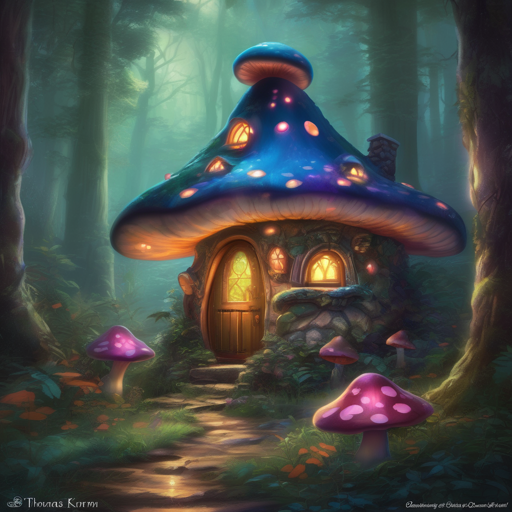}} \\ 
\midrule
\parbox[t]{\linewidth}{
\textbf{User}\,\,: Frog emerging from yogurt. \\ \\
\textbf{PRIP}: Frog emerging from yogurt., detailed, intricate, 4k, by Pauline Baynes, whimsical, award-winning, highly detailed, fantasy, magical, sparkle
} & 
\raisebox{-.85\height}{\includegraphics[width=\myimagewidth]{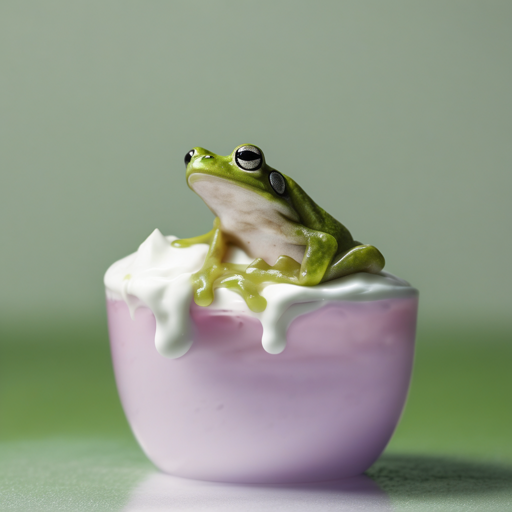}}  &
\raisebox{-.85\height}{\includegraphics[width=\myimagewidth]{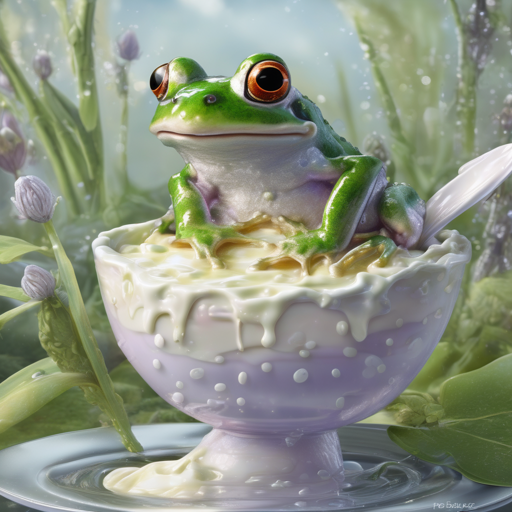}} \\ 
\midrule
\parbox[t]{\linewidth}{
\textbf{User}\,\,: A little cat is running in the woods \\ \\ \\
\textbf{PRIP}: A little cat is running in the woods, smiling, friendly, colorful, happy, laughing, cute, adorable
} & 
\raisebox{-.85\height}{\includegraphics[width=\myimagewidth]{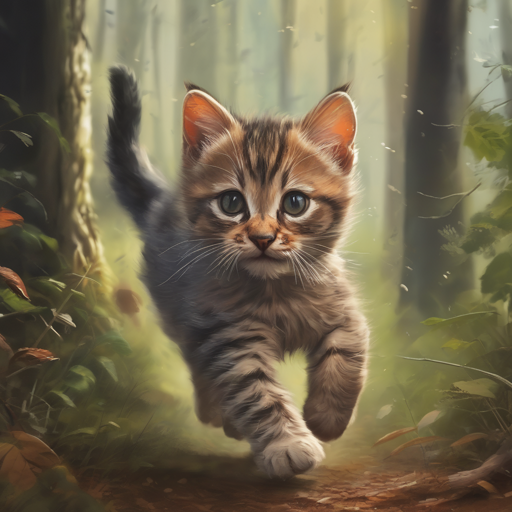}}  &
\raisebox{-.85\height}{\includegraphics[width=\myimagewidth]{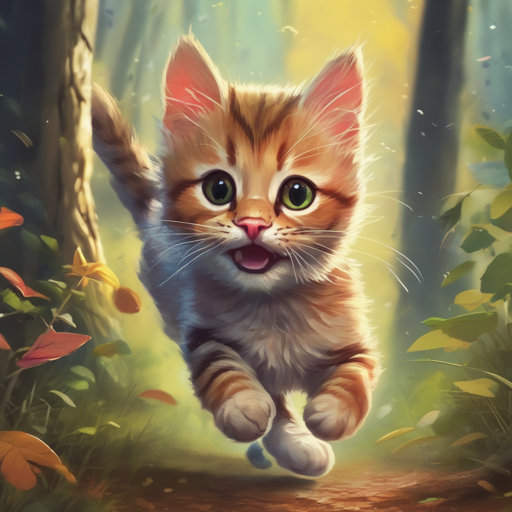}} \\ 
\midrule
\parbox[t]{\linewidth}{
%  A billboard posed by the side of a street in a rural town.
\textbf{User}\,\,: A billboard posed by the side of a street in a rural town. \\ 
\textbf{PRIP}: A billboard posed by the side of a street in a rural town., Christchurch, New Zealand, Hawaiian, small town, tropical, warm, fountain, happy, fun, touristy, cute, quaint % , main street, town square, summer
} & 
\raisebox{-.85\height}{\includegraphics[width=\myimagewidth]{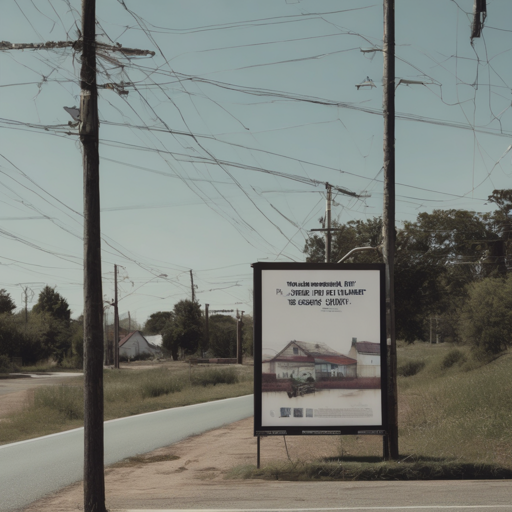}}  &
\raisebox{-.85\height}{\includegraphics[width=\myimagewidth]{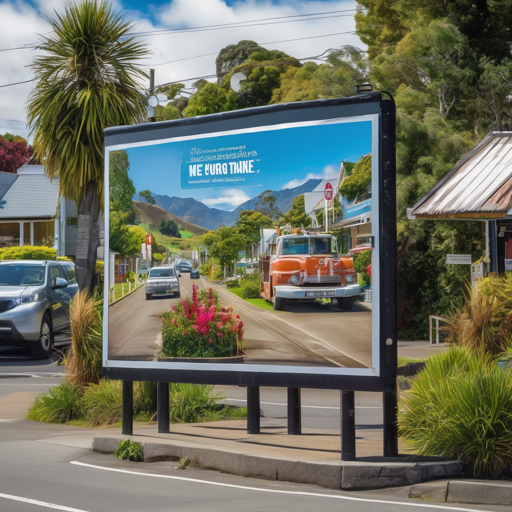}} \\ 
\midrule
\parbox[t]{\linewidth}{
\textbf{User}\,\,: A crowd of pink elephants playing steampunk instruments during a grindcore show. \\ 
\textbf{PRIP}: A crowd of pink elephants playing steampunk instruments during a grindcore show., cute, adorable, pastel colors, Thomas Kinkade, Colorful, Lisa Frank, family % friendly,
} & 
\raisebox{-.85\height}{\includegraphics[width=\myimagewidth]{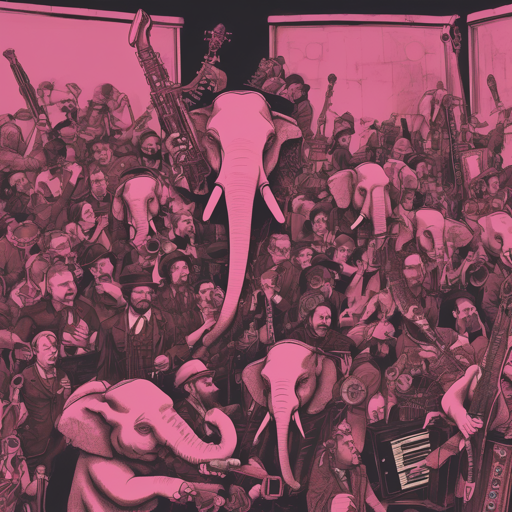}}  &
\raisebox{-.85\height}{\includegraphics[width=\myimagewidth]{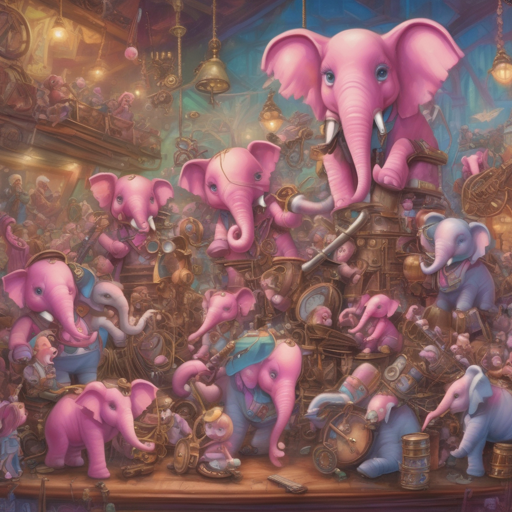}} \\ 
\midrule
%\parbox[t]{\linewidth}{
%\textbf{User}\,\,: A cute digital art of a unicorn. \\ 
%\textbf{PRIP}: A cute digital art of a unicorn., by Thomas Kinkade and Lisa Frank, intricate, colorful, highly detailed, artstation, ornate, luxury, beautiful, masterpiece, concept art, soft, sharp focus, illustration, art by artgerm, by Mark Ryden and Pixar
%} & 
%\raisebox{-.85\height}{\includegraphics[width=\myimagewidth]{./prompt_case_study/unicorn/sdxl.png}}  &
%\raisebox{-.85\height}{\includegraphics[width=\myimagewidth]{./prompt_case_study/unicorn/sdxl_prip.png}} \\ 
%\midrule
\parbox[t]{\linewidth}{
\textbf{User}\,\,: an empty bench sitting on the side of a sidewalk \\ 
\textbf{PRIP}: an empty bench sitting on the side of a sidewalk, in christchurch new zealand, small town, lots of flowers, small city, small town atmosphere, neighborhood, neighborhood atmosphere
} & 
\raisebox{-.85\height}{\includegraphics[width=\myimagewidth]{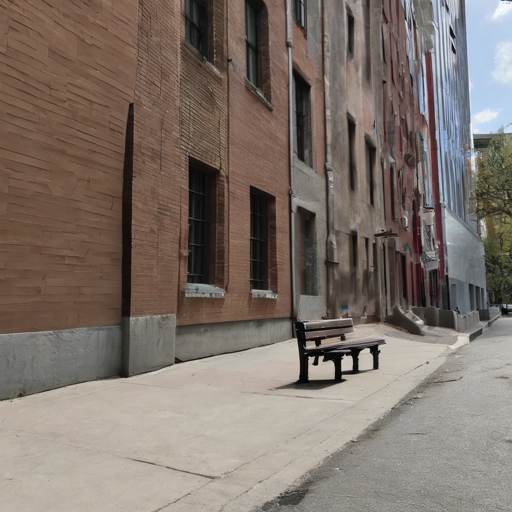}}  &
\raisebox{-.85\height}{\includegraphics[width=\myimagewidth]{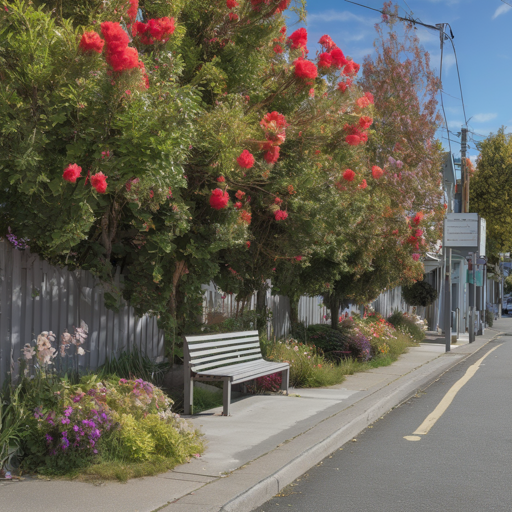}} \\ 
\bottomrule
\end{tabular}}
\caption{Refinement cases. Column one details user inputs and PRIP outputs. The next two columns show SDXL-generated images for different prompts. PRIP substantially enhances the image quality.}
\label{tab:prompt_case_study}
\end{table}

\let\myimagewidth\undefined

%% file: tables/ablation_study.tex
\newcommand{\B}{\bfseries}

\begin{table}[t]
\centering
\small
\scalebox{0.8}[0.8]{
\begin{tabular}{lcccc}
   	\toprule
%    \multirow{2}{*}{Method}
	Evaluation Metric & \multicolumn{2}{c}{ImageReward} & \multicolumn{2}{c}{HPSv2} \\ 
    Generation Model & SD1.4 & SDXL & SD1.4 & SDXL \\
    \midrule    
%    SD1.4 & 0.138 & 27.05 \\
    \multicolumn{3}{l}{\textit{without Reinforcement Learning}} \\
    PRIP\textbackslash RL & \B 0.122 & \B 0.888 & \B 27.24  & \B 27.87 	\\
    \;\textbackslash User-Pivot Preference & 0.072 & 0.862 & 27.17 & 27.80 \\
    \;\textbackslash Pivot-System Decoding & 0.058 & 0.810 & 27.04 & 27.69 \\
%    \;\textbackslash User-System Residual & -0.175 & 26.99 & 0.351 & 27.33 \\
    \midrule
    \multicolumn{3}{l}{\textit{with Reinforcement Learning}} \\
    PRIP & \B 0.404 & \B 0.983 & \B 27.77 & \B 28.15   	\\
    \;\textbackslash User-Pivot Preference & 0.198 & 0.917 & 27.33 & 27.92 \\
    \;\textbackslash Pivot-System Decoding & 0.047 & 0.821 & 27.04 & 27.71 \\
%    \;\textbackslash User-System Residual & 0.074 & 27.39 & 0.595 & 27.70 \\
    \bottomrule
\end{tabular}}
\caption{Ablation Study Results. The table presents the performance when various components of PRIP are removed. Results demonstrate their importance.}
\label{tab:ablation_study}
\end{table}

\let\B\undefined

%% file: tables/scaling.tex
\newcommand{\B}{\bfseries}
\newcommand{\sig}{$^{*}$}
\newcommand{\emp}{\,\;}

\begin{table}[t]
\centering
\small
\scalebox{0.8}[0.8]{
\begin{tabular}{lcccc}
   	\toprule
%    \multirow{2}{*}{Method}
	Evaluation Metric & \multicolumn{2}{c}{ImageReward} & \multicolumn{2}{c}{HPSv2} \\ 
    Dataset & Anime & Painting & Anime & Painting \\
    \midrule    
    SD1.4 & 0.038\emp & 0.190\emp & 27.42\emp & 26.89\emp \\
    \midrule
    \multicolumn{3}{l}{\textit{+ PRIP$\backslash$RL prompt refinement}} \\
    GPT2-Large 0.78B & 0.047\emp & 0.197\emp & 27.49\sig  & 27.00\sig 	\\
    TinyLlama 1.1B & 0.059\emp & 0.253\sig & 27.51\sig  & 26.99\sig 	\\
    Llama2 7B & \B 0.065\emp & \B 0.269\sig & \B 27.53\sig  & \B 27.03\sig 	\\
    \bottomrule
\end{tabular}}
\caption{PRIP performance when prompt decoder is initialized from different language models, including GPT2-large, TinyLlama, and Llama2. Results show that a larger model leads to better performance. $^*$ indicates the refinement model significantly outperforms the SD1.4 baseline (without refinement) with p-value $< 0.01$ measured by T-Test.}
\label{tab:scaling}
\end{table}

\let\B\undefined
\let\sig\undefined
\let\emp\undefined

%% file: sec/5_conclusion.tex
\section{Conclusion}

In this paper, we present PRIP, a pioneering pivot-based approach tailored for text-to-image prompt refinement. By formulating the refinement process as user-pivot preference encoding and pivot-system prompt decoding, PRIP sidesteps the scarcity of user-system refinement pairs and leverages large-scale data for effective model training. Extensive experiments demonstrate PRIP's effectiveness in prompt refinement. The improvement is pronounced for both text-to-image systems seen and unseen during training, highlighting its remarkable effectiveness and robust generalizability.

%% file: sec/ethics_impact.tex
\section{Limitation and Future Work}

There are several limitations for future studies:

(1) Dependence on Supervised Data: PRIP's Preference Encoder relies on image preference data to align with user preferences. This data requires manual annotation or user click logs. Future work should explore how to effectively train the Preference Encoder with minimal preference data.

(2) Dependence on System Language Corpus: PRIP's Prompt Decoder requires a corpus of system language prompts for training. Obtaining this data can rely on scraping prompts from websites or using interaction logs, which should be done with user consent and possibly compensation. Future work should investigate acquiring this data while protecting users' intellectual property and privacy.

(3) Usage of a Frozen Image Encoder: During PRIP's training, we use a frozen image encoder to encode images. Due to resource limitations, we did not explore the impact of different image encoders. Future work can explore how to select and train image encoders for PRIP.

(4) Transferability to New Systems: While PRIP shows promise in transferring to unseen systems, its long-term adaptability to rapidly evolving generation systems remains to be fully tested. Future work should investigate how to further improve PRIP's transferability, possibly by adapting the pivot-system module for different systems.

(5) Hallucination Problem: Since the Prompt Decoder cannot directly access user inputs when generating system language, hallucinations may occur. We address this issue by using the user input as a prefix during inference. Yet this restricts PRIP's capabilities, as using redundant or unclear user inputs as prefixes can affect the refinement effectiveness. Future work should further investigate this problem.
% to reduce hallucinations while enhancing PRIP's rewriting ability.

\section{Ethical Considerations}

The development and deployment of PRIP raise several ethical considerations that should be addressed in future applications:

(1) Intellectual Property Protection: PRIP relies on system languages provided by users for training. It is crucial to implement appropriate incentives and revenue-sharing mechanisms to protect the intellectual property of these users.

(2) Privacy Protection: When using prompt logs for training, user privacy must be fully respected. It is necessary to obtain users' consent in advance and implement appropriate anonymization measures.

(3) Bias and Fairness: Like all AI systems, PRIP's outputs are influenced by the data it is trained on. If image preference data contains biases, these biases can be reflected in the refined prompts, potentially showing societal stereotypes and biases in generated images.

(4) Misuse Potential: The enhanced capability in text-to-image generation can be misused for creating misleading or harmful content. Ensuring that PRIP and similar technologies are used responsibly requires robust guidelines and possibly technological safeguards against such misuse.

(5) Accessibility and Inclusion: By facilitating more intuitive interaction with text-to-image systems, PRIP contributes to making these technologies more accessible. However, it is crucial to ensure that these advancements are equally accessible to users across different languages, cultures, and socio-economic backgrounds.

\section{Broad Impact}

PRIP's development has broad implications for both society and the field of AI:

(1) Enhancing Creative Expression: By simplifying the process of prompt refinement, PRIP enables users, especially those without technical expertise, to more effectively leverage the power of text-to-image generation systems for creative expression, educational purposes, and more.

(2) Research in AI and Human-Computer Interaction: PRIP's novel approach to prompt refinement contributes to the understanding of how humans interact with AI systems. It opens new avenues for research in natural language processing, computer vision, and human-computer interaction, particularly in the context of improving the intuitiveness and effectiveness of AI interfaces.

(3) Potential Negative Consequences: While PRIP enhances user experience, the technology's ability to generate realistic images from refined prompts raises concerns about misinformation, privacy, and the ethical use of AI-generated content. It is important for the research community to address these issues and to develop ethical guidelines and best practices for the use of such technologies.

%% file: sec/appendix.tex
\section{Appendix}

\subsection{Comparison with ReFL}
\label{sec:compare_with_refl}

PRIP enhances input prompts for text-to-image generation systems, while alternative approaches like ReFL directly finetune the diffusion models themselves. Both utilize similar reward scores and reinforcement learning techniques. 

\input{tables/compare_to_refl}

\noindent \textbf{Quantitative Analysis: }
Table~\ref{tab:compare_to_refl} details a comparison of their performance. The automated metrics suggest that PRIP, despite only adjusting the input, rivals the performance of ReFL that finetunes the entire model. By feeding PRIP's refined prompts to ReFL during inference, we observe further improvements, implying that the strengths of the two methods are complementary. From the perspective of human evaluation, images generated by PRIP are notably preferred, a finding that diverges from the ImageReward scores. Considering that ReFL's training utilizes ImageReward, ReFL may adversarially attack this metric~\cite{akhtar2018threat}, resulting in seemingly high but perhaps not genuine scores. PRIP, while also trained with ImageReward, operates under strict constraints imposed by a fixed text-to-image model, thereby avoiding potential attack to the reward model.

%PRIP views generation systems as black boxes and only operates on the input while another model, ReFL, directly trains diffusion models. Both models use similar reward scores and RL processes. Table~\ref{tab:compare_to_refl} compares their performance. According to automated metrics, the performance of PRIP and ReFL is comparable, indicating that even though PRIP only manipulates the inputs, it can achieve similar performance to directly training diffusion models. We also combine the two approaches by feeding prompts rewritten by PRIP directly into ReFL during inference. The experimental results demonstrate further improvements, suggesting that their advantages are complementary. In terms of human evaluation metrics, we observe a strong preference for outputs generated by PRIP, contradicting the ImageReward scores. Since ImageReward is used for training ReFL, we hypothesize that directly training diffusion models makes ReFL ``hack'' ImageReward and results in inflated automated scores. Although ImageReward is also used to train PRIP, PRIP is strongly regularized with a frozen text-to-image generation system and thus avoids this issue.

\input{tables/refl_cases_no_sdxl}

\noindent \textbf{Image Examples: }
Table~\ref{tab:image_case_study} showcases the images generated by ReFL and PRIP. We can see that both ReFL and PRIP enhance the generation quality of SD1.4. For instance, the first and third examples illustrate that ReFL and PRIP align the generated images more closely with the prompts and that their combination achieves even better images. This observation is consistent with our quantitative findings. 

%Additionally, PRIP can be directly applied to SDXL in a zero-shot manner and boost its performance, whereas ReFL requires training on the generative model and cannot be transferred. SDXL+PRIP produces images that are more visually appealing than the rest. This highlights the advantage of prompt refinement in terms of generalization.

\subsection{Human Annotation Process}
\label{sec:detailed_annotation_instruction}

We recruit annotators to manually evaluate the output images from different generation systems. 
We recruit five annotators in total, from different backgrounds.
The annotator is paid for 30\$ per hour. Each data item is labeled by three annotators, and the median value is used as the final label.

The annotators are informed about the intended usage of the data: "The data annotated is used for scientific research, and does not involve any personal privacy. The annotation process will not have any physical or psychological impact on the subjects. The results of this research may be published in academic conferences/journals/books, or used for teaching. The dataset may be made public, but it is only for research by the academic community, and your name or other information that may identify you will not appear in any published or teaching materials."

We provide the instructions for the Preference Annotation task and the Relevance Annotation Task in the following.

\subsubsection{Preference Annotation}

The instruction for the Preference Annotation task is mainly the same as \citet{wu2023humanv2}:
``
We will provide a prompt that describes the image the user wants to draw. You will see two images, which are generated from two different AI models. Please consider the prompt and choose the better image from the perspectives of universal and personal aesthetic appeal. 
This task mainly involves two aspects: text-image alignment and image quality. Although we encourage and value personal preference, it’s important to consider the following fundamental principles when balancing the two aspects or facing a dilemma:
(1) When Image (A) surpasses Image (B) in terms of aesthetic appeal and fidelity, or Image (B) suffers from severe distortion and blurriness, if Image (B) aligns only slightly better with the prompt, Image (A) should take precedence over Image (B).
(2) When facing a dilemma that images are relatively similar in terms of aesthetics and personal preference, please carefully read and consider the prompt for sorting based more on the text-image alignment.
(3) It is crucial to pay special attention to the capitalized names. If there is any term or content you are not familiar with, we recommend you to search for sample images and explanations online.
''

\subsubsection{Relevance Annotation}

We will provide a prompt that describes the image the user wants to draw. You will see one image, which is generated from an AI system. Please consider the prompt and choose how relevant the image is to the prompt. 
\begin{itemize}
	\item 0: The image is completely irrelevant to the given prompt. The image does not contain any of the key entities mentioned in the prompt. This also applies if the image only matches the prompt in style or detail, but does not contain the corresponding key entities. 
	\item 1: The image is partially relevant to the given prompt. The image contains some of the key entities mentioned in the prompt, but may differ in style, action, or detail.
	\item 2: The image is perfectly relevant to the given prompt. The subject, style, and details of the image are all consistent with the prompt.
\end{itemize}

\subsection{Datasheet}

In this paper, we use two public datasets, namely DiffusionDB~\cite{wang2022diffusiondb} for training and HPSv2 benchmark~\cite{wu2023humanv2} for evaluation.
\begin{itemize}
	\item License: The DiffusionDB dataset is under ``CC0 1.0 License'' license. The HPSv2 benchmark is under `Apache-2.0 license'.
	\item Intended use: our use is consistent with the dataset creators' intended use. For DiffusionDB, the authors stated that the dataset can help ``design human-AI interaction tools to help users more easily use these models.''. The HPSv2 benchmark is exactly proposed to evaluate text-to-image generation performance.
	\item Content Processing: Since we directly use the two public datasets and do not preprocess the datasets, we also inevitably use the potentially harmful content from the two datasets. According to both creators, the data is filtered by NSFW classifiers but still may contain a small portion of harmful content.
	\item Coverage: The two datasets cover a wide range of topics, including anime, concept art, paintings, and realistic photos. The languages are mainly English, and also include other languages like Japanese and Chinese.
	\item Train/Val/Test: We use DiffusionDB for training and validation. We construct 300k image preference pairs and 900k system language prompts for training. We randomly sample 1,000 prompts from DiffusionDB for validation. HPSv2 is used as the test set. It contains four categories, each consisting of 800 prompts.
\end{itemize}

\subsection{Computational Experiments}
\label{sec:computational_exp}
Preference Encoder is initialized from Flan-T5-Large~\cite{chung2022scaling} and is of 738M parameters. Prompt Decoder is initialized from Llama 2 7B~\cite{touvron2023llama} and is of 7B parameters. We use Transformers library~\cite{wolf-etal-2020-transformers} for training and inference. User-pivot warmup, pivot-system warmup and RL takes $24$, $144$, and $384$ GPU hours on A100 devices.

We empirically tune the training hyper-parameters such as learning rate to minimize the validation loss on a held-out set. 
During warmup, Preference Encoder is trained for $3$ epochs with a learning rate of $0.001$, and Prompt decoder is trained for $2$ epochs and a learning rate of $2\times 10^{-5}$.
During user-pivot-system RL training, we use ImageReward and HPSv2 to output preference scores, and train PRIP for $1,000$ steps with a batch size of $512$ and a constant learning rate of $0.001$

\subsection{PRIP Model Card}
\label{sec:model_card}

The two components of PRIP, namely the Preference Encoder and the Prompt Decoder, share the same model architecture with Flan-T5-Large~\cite{chung2022scaling} and Llama 2~\cite{touvron2023llama}, respectively.

\begin{table}[h]
\small
\centering
\setlength{\tabcolsep}{5mm}
\begin{tabular}{ll}
   	\toprule
   	\multicolumn{2}{c}{Preference Encoder} \\
   	Initialization & Flan-T5-Large  \\
   	Input & Text \\
   	Output & $\mathbb{R}^{32 \times 768}$ \\
   	\midrule 
   	\multicolumn{2}{c}{Prompt Decoder} \\
   	Initialization & Llama 2  \\
   	Input & $\mathbb{R}^{32 \times 768}$ \\
   	Output & Text \\
   	\bottomrule
\end{tabular}
\caption{PRIP Model Card.}
\end{table}

\subsection{Use of AI Assistant}

We used AI assistant tools such as ChatGPT for polishing. However, AI-generated text is only used for reference in writing and is added to the article after careful consideration and modification. The help of AI lies in providing suggestions to make the paper more readable. We do not directly copy large chunks of text generated by ChatGPT into our paper without checking or modification.

%% file: tables/compare_to_refl.tex
\newcommand{\B}{\bfseries}

\begin{table}[h]
\centering
\small
\setlength{\tabcolsep}{1.8mm}
\scalebox{0.8}[0.8]{
\begin{tabular}{lccccc}
   	\toprule
    \multirow{2}{*}{Method} & \multicolumn{5}{c}{Average on Four Datasets} \\
    & ImageReward & HPSv2 & Relevance & Win & Win+Tie \\
    \midrule
ReFL           & 0.421  & 27.64 & 1.42 & 0\% & \B 100\% \\
SD1.4+\textbf{PRIP}            & 0.404  & 27.77 & \B 1.45 & 53\% & 80\% \\
ReFL+\textbf{PRIP}       & \B 0.640   & \B 28.14 & 1.44 & \B 64\% & 89\% \\
    \bottomrule
\end{tabular}}
\caption{Comparison with ReFL, a SD1.4 model finetuned with user preference. Win ratio shows preference against ReFL. Results highlight that PRIP outperforms ReFL in manual evaluations and that integrating PRIP with ReFL yields further enhancements. }
\label{tab:compare_to_refl}
\end{table}

\let\B\undefined

%% file: tables/refl_cases_no_sdxl.tex
\newlength{\myimagewidth}
\setlength{\myimagewidth}{0.18\linewidth} % 设置您希望的全局宽度

\begin{table}[h]
\centering
\scriptsize
\setlength{\tabcolsep}{1pt}
\scalebox{0.9}[0.9]{
\begin{tabular}{m{0.28\linewidth}cccc}
\toprule
\textbf{User Prompt} & \textbf{SD1.4} & \textbf{ReFL} & \textbf{SD1.4+PRIP} & \textbf{ReFL+PRIP}  \\ 
\midrule
A digital art depicting a chicken wearing a suit. & 
\raisebox{-.5\height}{\includegraphics[width=\myimagewidth]{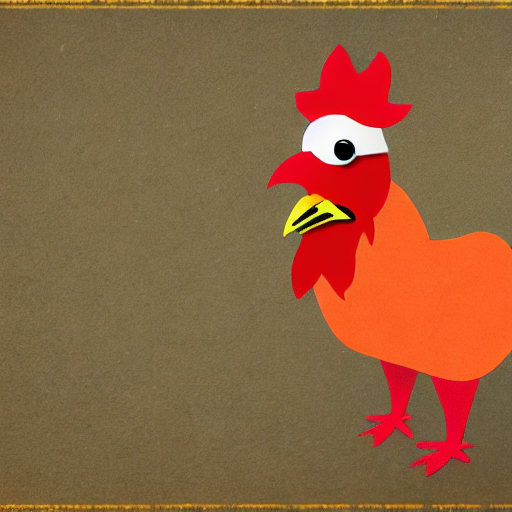}} & 
\raisebox{-.5\height}{\includegraphics[width=\myimagewidth]{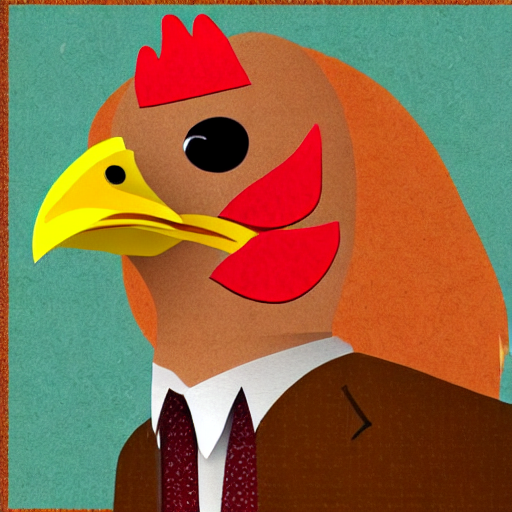}} & 
\raisebox{-.5\height}{\includegraphics[width=\myimagewidth]{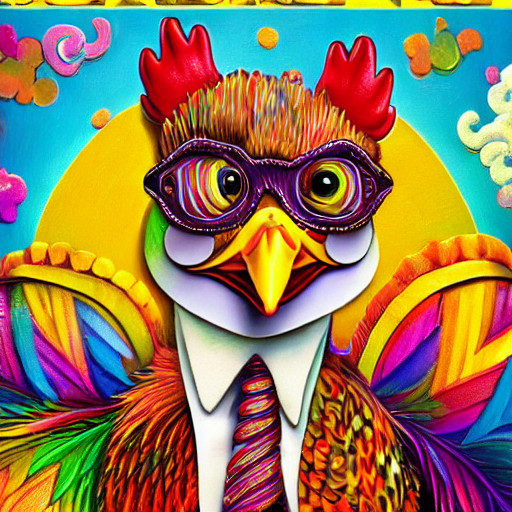}} &
\raisebox{-.5\height}{\includegraphics[width=\myimagewidth]{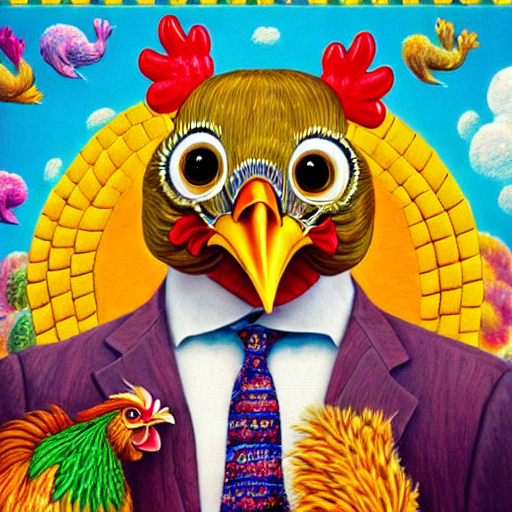}} 
\\
\midrule 
The interior of a spaceship orbiting alpha centauri. &
\raisebox{-.5\height}{\includegraphics[width=\myimagewidth]{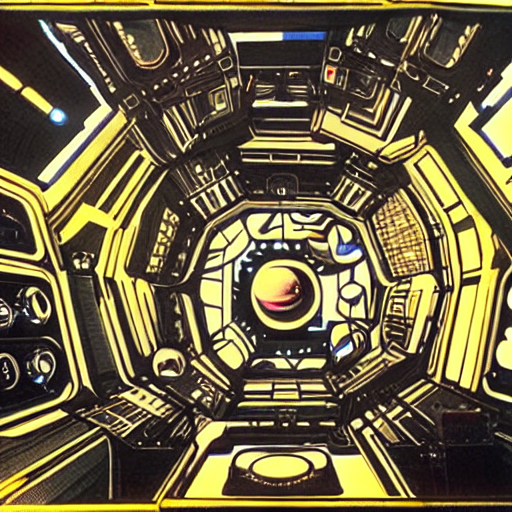}} & 
\raisebox{-.5\height}{\includegraphics[width=\myimagewidth]{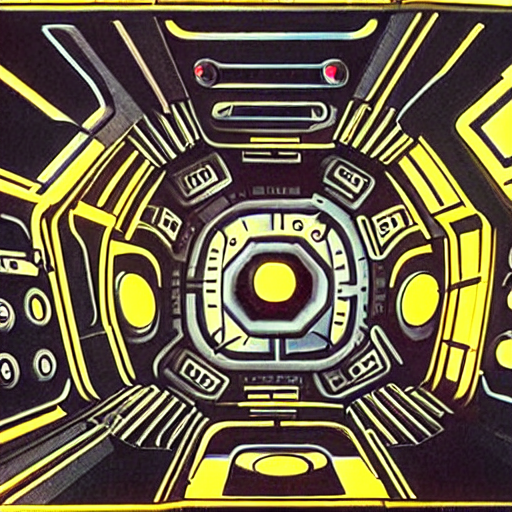}} & 
\raisebox{-.5\height}{\includegraphics[width=\myimagewidth]{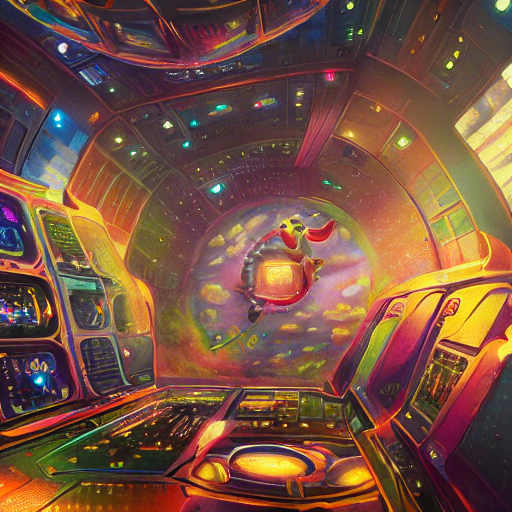}} &
\raisebox{-.5\height}{\includegraphics[width=\myimagewidth]{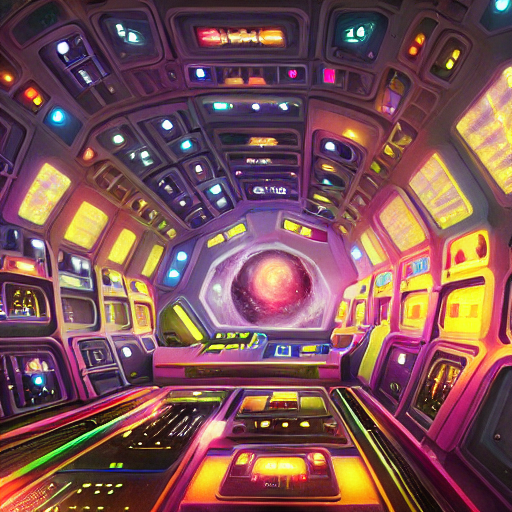}} 
\\
\midrule
A horse and astronaut in one image. &
\raisebox{-.5\height}{\includegraphics[width=\myimagewidth]{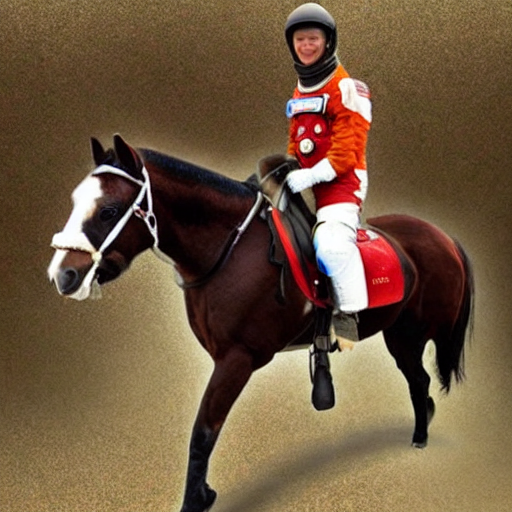}} & 
\raisebox{-.5\height}{\includegraphics[width=\myimagewidth]{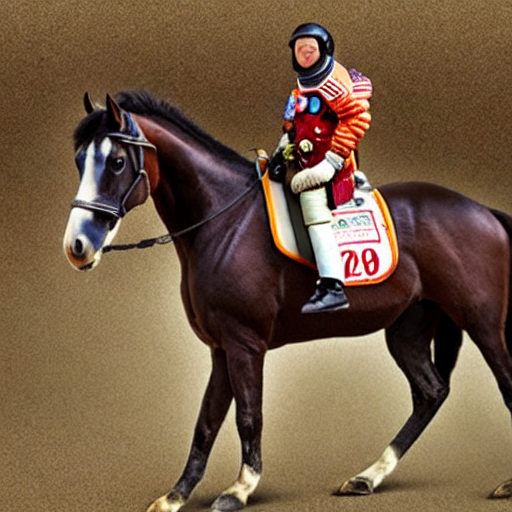}} & 
\raisebox{-.5\height}{\includegraphics[width=\myimagewidth]{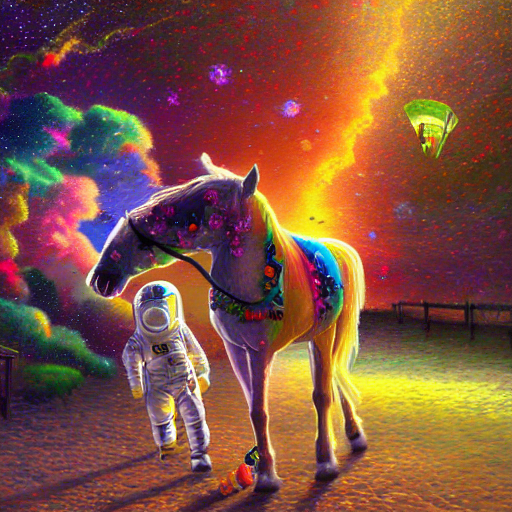}} &
\raisebox{-.5\height}{\includegraphics[width=\myimagewidth]{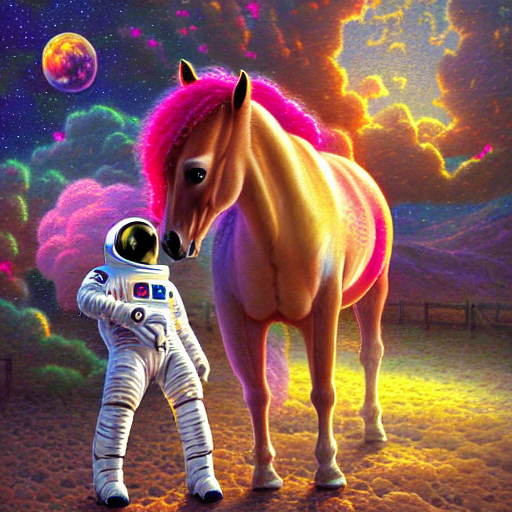}} 
\\
\midrule
A plane flies in the sky passing over the moon. &
\raisebox{-.5\height}{\includegraphics[width=\myimagewidth]{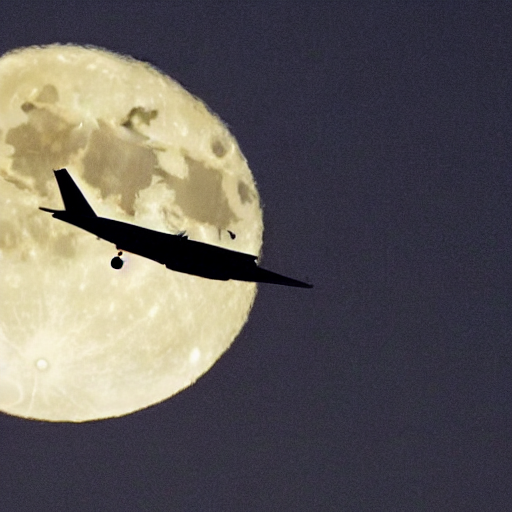}} & 
\raisebox{-.5\height}{\includegraphics[width=\myimagewidth]{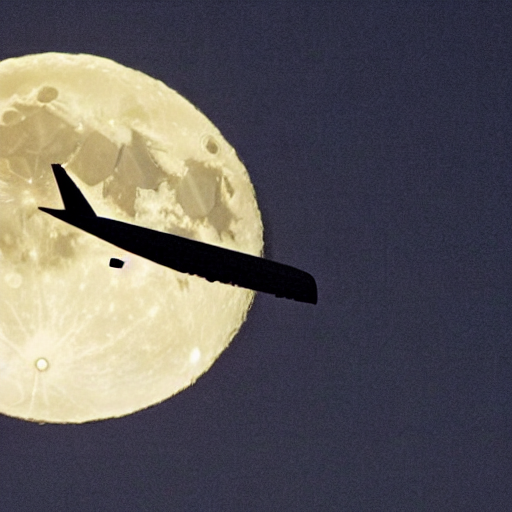}} & 
\raisebox{-.5\height}{\includegraphics[width=\myimagewidth]{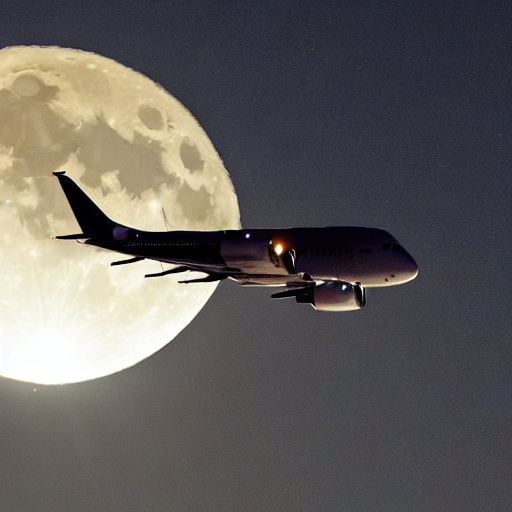}} &
\raisebox{-.5\height}{\includegraphics[width=\myimagewidth]{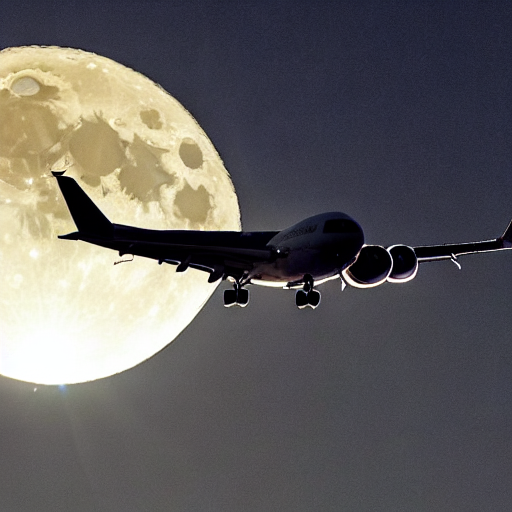}} 
\\
\bottomrule
\end{tabular}}
\caption{Examples of PRIP and ReFL: The table showcases the user prompts in the first column, followed by images generated using different methods in the subsequent columns. Both ReFL and PRIP enhance the performance of SD1.4 individually, and their combination yields even better outcomes. \label{tab:image_case_study}}
\end{table}

\let\myimagewidth\undefined